\documentclass[10pt,a4paper]{article}
\usepackage[hmargin=2cm,vmargin=2.2cm]{geometry}
\usepackage[T1]{fontenc}
\usepackage{lmodern}
\usepackage{xcolor}

\usepackage[numbers,sort&compress]{natbib}
\usepackage{amsmath}
\usepackage{amssymb}
\usepackage{amsthm}
\usepackage{mathtools}
\usepackage{enumitem}
\usepackage{graphicx}
\usepackage{tabularx}
\usepackage{array}
\usepackage{multirow}
\usepackage{algorithm}
\usepackage{algpseudocode}
\usepackage{url}
\usepackage{listings}
\newcolumntype{Y}{>{\raggedright\arraybackslash}X}
\newcolumntype{L}[1]{>{\raggedright\arraybackslash}p{#1}}

\newcommand{\figorplaceholder}[2]{\IfFileExists{#1}{\includegraphics[width=#2]{#1}}{\fbox{\parbox[c][0.25\textheight][c]{0.9#2}{\centering Figure file \texttt{\detokenize{#1}} not yet rendered}}}}

\newtheorem{theorem}{Theorem}
\newtheorem{lemma}[theorem]{Lemma}
\newtheorem{proposition}[theorem]{Proposition}
\newtheorem{corollary}[theorem]{Corollary}
\newtheorem{fact}[theorem]{Fact}
\newtheorem{conjecture}[theorem]{Conjecture}
\theoremstyle{definition}
\newtheorem{definition}[theorem]{Definition}
\newtheorem{remark}[theorem]{Remark}

\newcommand{\Ldim}{\operatorname{Ldim}}
\newcommand{\Int}{\mathrm{Int}}
\newcommand{\lgf}[1]{\lambda(#1)}
\newcommand{\N}{\mathbb{N}}
\usepackage{xurl}
\usepackage[colorlinks=true,allcolors=blue]{hyperref}

\begin{document}
\setlength{\emergencystretch}{2em}
\title{How many mistakes does it take to learn an interval? Exact Littlestone dimensions of intervals and their unions}
\author{Miit Daga\thanks{ORCID: 0009-0005-4629-458X. Email: \href{mailto:miitdaga03@gmail.com}{miitdaga03@gmail.com}}
\qquad Swarna Priya Ramu\thanks{Corresponding author. ORCID: 0000-0002-8287-9690. Email: \href{mailto:swarnapriya.rm@vit.ac.in}{swarnapriya.rm@vit.ac.in}}\\[6pt]
\normalsize School of Computer Science Engineering and Information Systems,\\
\normalsize Vellore Institute of Technology, Vellore 632014, Tamil Nadu, India}
\date{}
\maketitle

\begin{abstract}
The Littlestone dimension of a hypothesis class equals the smallest number of mistakes an online learner can guarantee against an adaptive adversary. For discrete intervals and unions of $j$ intervals on $m$ points, only its order of growth was known. We determine it exactly. For every switch budget $k$ (with both end labels fixed when $k$ is odd), the smallest size on which $d$ mistakes can be forced follows a three-regime law: $d+1$ while $d\le k$, then $2d-k+1$, and beyond $d=\lceil 3k/2\rceil+k-1$ doubling the size adds exactly $k$ mistakes. Unions of $j$ intervals on $m$ points correspond to $k=2j$ and size $m+1$, and a single interval gives the thresholds $1,2,3,5,7,10,14,20,\dots$. The natural learner that waits for a positive example and binary-searches both endpoints is not optimal: it can be forced into one extra mistake for $50{,}847$ of the domain sizes below $10^5$. Upper bounds come from a gap potential and a block-pairing learner, and lower bounds from adversary strategies and a refinement step. These tools give an explicit optimal learner with $O(\log m)$ time per round, which on random streams makes more mistakes on average than the natural learner. For randomised learners, we compute the randomised Littlestone dimension of thresholds exactly and prove that a greedy adversary forces an explicit closed-form number of expected mistakes on intervals; this bound is within $0.64$ of optimal and, we conjecture, exact (confirmed for $m\le70$). Exact computation tests every result, and the formulas for every switch budget are machine-checked in Lean~4.
\end{abstract}

\noindent\textbf{Keywords:} Online learning; Littlestone dimension; Mistake bounds; Interval classes; Randomised learners

\section{Introduction}

\begin{figure}[!htbp]
\centering
\includegraphics[width=\textwidth]{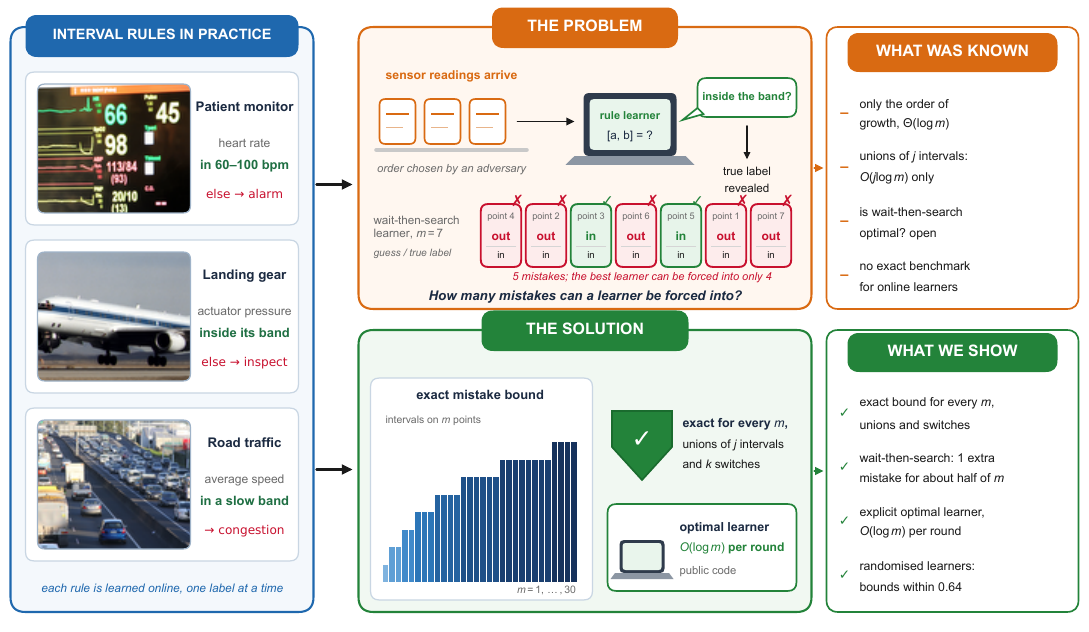}
\caption{From interval rules to exact mistake bounds. Left: examples of interval rules (the settings shown are illustrative), each learned online from one label at a time. Top: the online game. The strip shows a worst-case game against the natural wait-then-search learner on $m=7$ points, with the learner's guess above the true label on each card; the learner makes $5$ mistakes, one more than $\Ldim(\Int_7)=4$. Bottom: the results; the bars are $\Ldim(\Int_m)$ for $m\le30$. Right: what was known before and what this paper shows. Photographs, all from Wikimedia Commons: patient monitor by J.~Stenberg, U.S.\ Navy (public domain); aircraft by T.~Landis, NASA (public domain); road traffic by Kiwiev (CC0~1.0).}
\label{fig:motivation}
\end{figure}

Many knowledge-based and decision-support systems state their rules as intervals on an ordered attribute, for example the condition-monitoring systems that check aircraft landing-gear actuators against fault thresholds \cite{phillips2011knowledge}. An alarm is raised when a reading falls inside a band, and a classification rule fires when a feature lies between two cut points; a monitoring rule may also change state a few times along a time or measurement axis. Such rules are often learned online from labels that arrive one at a time. Hoeffding trees and incremental model trees learn threshold splits from data streams \cite{domingos2000mining,wibisono2016traffic}, and stream rule learners and evolving fuzzy classifiers update their rules one example at a time \cite{kosina2015very,duarte2016adaptive,lughofer2022evolving}. When the labels may also come from an adversarial source, the natural guarantee is the number of mistakes a learner can be forced into before the rule is identified. For intervals, unions of intervals and binary sequences with few switches, this number has so far been known only up to constant factors. Figure~\ref{fig:motivation} sketches this setting and the results of this paper.

Littlestone \cite{littlestone1988learning} showed that the optimal mistake bound of a concept class $H$ in realisable online learning equals a combinatorial parameter, now called the Littlestone dimension $\Ldim(H)$, and that the Standard Optimal Algorithm attains it. Exact values of $\Ldim$ are known for only a handful of natural classes, and computing the dimension of an explicitly given class is hard in general \cite{frances1998optimal,manurangsi2017inapproximability}. For the half-intervals $\{1,\dots,i\}$ on $\{1,\dots,n\}$, Maass and Tur\'an \cite[Prop.~4.2]{maass1992lower} proved that the optimal mistake bound with arbitrary hypotheses is exactly $\lfloor\lg n\rfloor$, i.e.\ binary search is optimal; more generally a chain of $c$ threshold functions has dimension $\lfloor\lg c\rfloor$ (Lemma~\ref{lem:thr}). For axis-parallel boxes over $\{1,\dots,n\}^d$ the picture is asymptotic: Maass and Tur\'an \cite{maass1994algorithms} showed that the proper learning complexity is $\Theta(\log n)$ for every fixed $d$, and Chen and Maass \cite{chen1992line,chen1994line} showed that the optimal mistake bound with arbitrary hypotheses is $\Theta(d\log n)$. Recent work states the bounds for intervals and unions of $j$ intervals only as $O(j\log T)$ on $T$ points \cite{attias2026tradeoffs}, and a 2026 study of autoregressive online learning lists sharper mistake bounds for ``finite-domain geometric classes such as intervals and unions of intervals'' as an open direction \cite{doron2026theory}. To our knowledge, no exact value has been published for intervals or for unions of intervals.

\begin{figure}[!htbp]
\centering
\includegraphics[width=\textwidth]{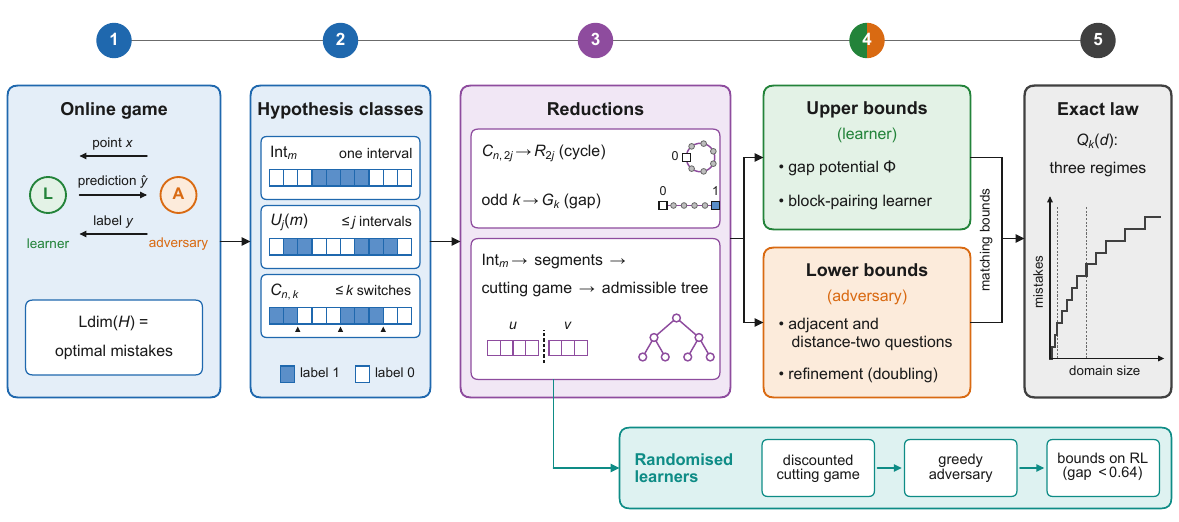}
\caption{Structure of the proofs. The online mistake-bound game (1) is posed for intervals, unions of intervals and sequences with few switches (2). Reductions (3) turn sequences with $2j$ switches into the cycle family $R_{2j}$, odd switch budgets into the gap family $G_k$, and intervals into a cutting game whose adversary strategies are admissible trees. Learner arguments give upper bounds and adversary arguments give lower bounds (4); the two meet in the three-regime law for $Q_k(d)$ (5). The branch from the cutting game treats randomised learners and gives lower and upper bounds on the randomised Littlestone dimension that differ by less than $0.64$.}
\label{fig:framework}
\end{figure}

We determine the one-dimensional case exactly. Learning an interval looks like two binary searches, about $2\lg m$ mistakes, but the exact answer has odd constants: for $m\ge3$ the dimension is $2\lg(m+1)$ minus a quantity that oscillates between $2\lg 7-4\approx1.615$ and $2\lg5-2\approx2.644$. The constants $5$ and $7$ come from the interaction between the first positive example and the two binary searches that follow it. A consequence is that the natural learner, which waits for a positive example and then searches for both endpoints, is not optimal: it can be forced into one extra mistake for about half of the domain sizes below $10^5$, and never into more than one. For unions of $j$ intervals the same mechanism produces a three-regime formula, and once $m+1\ge4j+1$, doubling $m+1$ adds exactly $2j$ mistakes, one per endpoint. An exact value is also a benchmark: an online learner for these classes can be measured against the optimal mistake bound itself rather than against an asymptotic estimate. Figure~\ref{fig:framework} shows how the proofs fit together.

\paragraph{Notation.} Throughout, $[m]=\{1,\dots,m\}$ (with $[0]=\emptyset$, so $\Int_0=\{\emptyset\}$) and $\lg=\log_2$, and for an integer $x\ge 1$ we write $\lgf{x}=\lfloor \lg x\rfloor$. A \emph{switch} of $f:[n]\to\{0,1\}$ is an index $i<n$ with $f(i)\ne f(i+1)$. We study three classes, each viewed as a class of indicator functions.
\begin{itemize}[leftmargin=*]
\item \emph{Intervals} on $[m]$:
\[
\Int_m=\{\emptyset\}\cup\bigl\{\{a,a+1,\dots,b\}:\ 1\le a\le b\le m\bigr\}.
\]
\item \emph{Unions of intervals}: for $j\ge1$, $U_j(m)$ is the class of subsets of $[m]$ that are unions of at most $j$ intervals, so $U_1(m)=\Int_m$.
\item \emph{Switch-bounded sequences}: for $n\ge1$ and $k\ge0$, $C_{n,k}$ is the class of all $f:[n]\to\{0,1\}$ with at most $k$ switches.
\end{itemize}

\paragraph{Main results.} The answer for intervals is governed by the sequence
\[
P(d)=\begin{cases} d+1, & 0\le d\le 2,\\ 5\cdot 2^{t-1}, & d=2t+1\ge3,\\ 7\cdot 2^{t-2}, & d=2t\ge4,\end{cases}
\]
which satisfies $P(d)=2P(d-2)$ for $d\ge 5$. Its first values are
\[
\setlength{\arraycolsep}{3.5pt}
\begin{array}{|c|c|c|c|c|c|c|c|c|c|c|c|c|}
\hline
d & 0&1&2&3&4&5&6&7&8&9&10&11\\ \hline
P(d) & 1&2&3&5&7&10&14&20&28&40&56&80\\ \hline
\end{array}
\]

\begin{theorem}\label{thm:main}
For every $m\ge 0$,
\[
\Ldim(\Int_m)=\max\{d\ge0:\ P(d)\le m+1\}.
\]
Equivalently, for $m\ge 3$,
\[
\Ldim(\Int_m)=\max\bigl(A_m,\,B_m\bigr),
\]
where
\[
A_m=2\Bigl\lfloor \lg\frac{4(m+1)}{7}\Bigr\rfloor,\qquad B_m=2\Bigl\lfloor \lg\frac{2(m+1)}{5}\Bigr\rfloor+1 .
\]
\end{theorem}

\begin{theorem}\label{thm:switch}
For every $n\ge 1$ and $j\ge 1$, with $U_j(0)=\{\emptyset\}$,
\[
\Ldim(C_{n,2j})=1+\Ldim\bigl(U_j(n-1)\bigr).
\]
In particular,
\[
\Ldim(C_{n,2})=1+\max\{d\ge 0:\ P(d)\le n\}.
\]
\end{theorem}

\begin{corollary}\label{cor:asym}
For all $m\ge 3$,
\[
\begin{aligned}
\Ldim(\Int_m)&>2\lg(m+1)-(2\lg 5-2),\\
\Ldim(\Int_m)&\le2\lg(m+1)-(2\lg 7-4).
\end{aligned}
\]
The upper bound is attained for infinitely many $m$, namely at $m+1=7\cdot 2^{t-2}$ with $t\ge2$. The lower bound is never attained, but it is approached arbitrarily closely from above at $m+1=5\cdot2^{t-1}-1$ as $t\to\infty$.
\end{corollary}

\begin{figure}[!htbp]
\centering
\includegraphics[width=\textwidth]{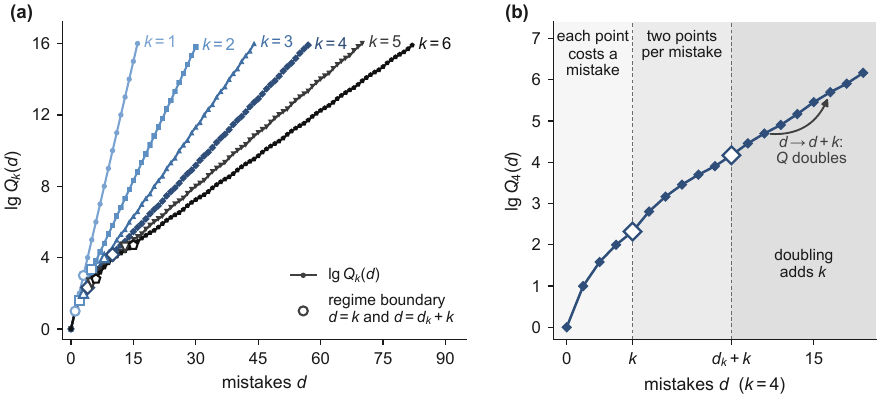}
\caption{The thresholds $Q_k(d)$, the smallest size on which $d$ mistakes can be forced. (a) $\lg Q_k(d)$ for $k=1,\dots,6$, plotted while $Q_k(d)\le2^{16}$; open markers show the regime boundaries $d=k$ and $d=d_k+k$, with $d_k=\lceil3k/2\rceil$. (b) The case $k=4$. Up to $d=k$ each added point gives one more mistake, up to $d=d_k+k-1$ every two added points give one, and from $d=d_k+k$ on, $k$ more mistakes need a domain twice as large.}
\label{fig:regimes}
\end{figure}

\paragraph{Arbitrary switch budgets.} For $k\ge1$ let $d_k=\lceil 3k/2\rceil$ and define $Q_k:\N\to\N$ by
\[
Q_k(d)=\begin{cases} d+1, & 0\le d\le k,\\ 2d-k+1, & k\le d\le d_k+k-1,\\ 2\,Q_k(d-k), & d\ge d_k+k,\end{cases}
\]
\[
v_k(x)=\max\{d\ge0:Q_k(d)\le x\}.
\]
Three facts help to read this definition.
\begin{itemize}[leftmargin=*]
\item The first two lines agree at $d=k$.
\item For $k=1$ it gives $Q_1(d)=2^d$ and $v_1(x)=\lgf{x}$; for $k=2$ it gives $Q_2=P$.
\item For $k=2j$ and $d=3j+2jt+i$ with $t\ge0$ and $0\le i<2j$,
\[
Q_{2j}(d)=(4j+2i+1)\,2^t .
\]
\end{itemize}
The first values for $k\le4$ are
\[
\setlength{\arraycolsep}{2.5pt}
\begin{array}{|c|c|c|c|c|c|c|c|c|c|c|c|c|}
\hline
d & 0&1&2&3&4&5&6&7&8&9&10&11\\ \hline
Q_1(d) & 1&2&4&8&16&32&64&128&256&512&1024&2048\\ \hline
Q_2(d) & 1&2&3&5&7&10&14&20&28&40&56&80\\ \hline
Q_3(d) & 1&2&3&4&6&8&10&12&16&20&24&32\\ \hline
Q_4(d) & 1&2&3&4&5&7&9&11&13&15&18&22\\ \hline
\end{array}
\]
Figure~\ref{fig:regimes} plots $\lg Q_k(d)$ for $k\le6$ and marks the three regimes.

Two families carry the general result; in both, $x$ is the \emph{size}.
\begin{itemize}[leftmargin=*]
\item \emph{Cycle family}, $k=2j$: $R_k(x)$ is the class of functions on a cycle of $x$ points, one of which is fixed to label $0$, with at most $k$ switches around the cycle. By Lemma~\ref{lem:cyc}, $R_{2j}(m+1)\cong U_j(m)$.
\item \emph{Gap family}, $k=2s+1$: $G_k(x)$ is the class of binary sequences $f(0),f(1),\dots,f(x)$ with $f(0)=0$ and $f(x)=1$ fixed and at most $k$ switches, viewed as a class on the $x-1$ free points.
\end{itemize}

\begin{theorem}\label{thm:general}
For every $k\ge1$ and $x\ge1$,
\[
\begin{aligned}
\Ldim\bigl(R_k(x)\bigr)&=v_k(x)\qquad(k\ \text{even}),\\
\Ldim\bigl(G_k(x)\bigr)&=v_k(x)\qquad(k\ \text{odd}).
\end{aligned}
\]
\end{theorem}

\begin{corollary}\label{cor:unions}
For all $j\ge1$, $m\ge0$ and $n\ge1$,
\[
\begin{aligned}
\Ldim(U_j(m))&=v_{2j}(m+1),\\
\Ldim(C_{n,2j})&=1+v_{2j}(n).
\end{aligned}
\]
If $m+1\ge4j+1$, then
\[
\begin{aligned}
\Ldim(U_j(m))&>2j\lg\frac{m+1}{j}-1.173\,j-1.722,\\
\Ldim(U_j(m))&\le2j\lg\frac{m+1}{j}-j .
\end{aligned}
\]
\end{corollary}

\begin{proposition}\label{prop:odd}
For all $n\ge1$ and $s\ge0$, with $k=2s+1$,
\[
\begin{aligned}
\Ldim(C_{n,k})&\ge1+v_k(n),\\
\Ldim(C_{n,k})&\le1+\min\bigl(\lfloor (n+k-1)/2\rfloor,\,v_{k+1}(n)\bigr).
\end{aligned}
\]
For $k=1$ the lower bound is attained:
\[
\Ldim(C_{n,1})=1+\lgf{n}.
\]
\end{proposition}

We conjecture that $\Ldim(C_{n,k})=1+v_k(n)$ for all odd $k$; this holds for $k=1$ and was checked by exhaustive search for $k=3$ ($n\le26$), $k=5$ ($n\le18$) and $k=7$ ($n\le15$). 

Section~\ref{sec:related} places these results in the literature. Sections~\ref{sec:prelim} to~\ref{sec:tree} prove Theorem~\ref{thm:main} directly. After the first query the version space splits into segments, the game becomes a cutting game on a rod (Section~\ref{sec:cut}), and the cutting game is solved by minimising the root measure of heap-labelled admissible trees (Section~\ref{sec:tree}). Section~\ref{sec:general} proves Theorem~\ref{thm:general} with a gap potential, a block-pairing learner and a small set of adversary moves; for $k=2$ it gives a second, shorter proof of Theorem~\ref{thm:main}. We keep the first proof for two reasons. Section~\ref{sec:random} reuses the cutting game for randomised learners, where the depth constraint turns into a discount. The admissible trees describe every winning cut sequence of the adversary, and the quantity $c(r)$ of Section~\ref{sec:tree} gives the exact cost of the natural learner (Remark~\ref{rem:natural}). The consequences for deterministic learners, including the optimal learner that the upper-bound proof of Section~\ref{sec:general} defines and its behaviour on random streams, are in Section~\ref{sec:remarks}. Section~\ref{sec:verification} lists the exact computations used as an independent check, and Section~\ref{sec:open} concludes with limitations and open problems.

\section{Related work}\label{sec:related}

Table~\ref{tab:comparison} positions this study relative to existing work on mistake bounds for geometric classes, exact optimal values for concrete classes, the computation of the Littlestone dimension, and randomised, transductive and self-directed variants of online learning.

\begin{table}[!htbp]
\centering
\caption{Prior work most comparable to this study: setting, key finding, what is evaluated, and the specific gap this study addresses relative to that work. The final row summarises this study.}
\label{tab:comparison}
\footnotesize\setlength{\tabcolsep}{3pt}
\begin{tabularx}{\textwidth}{|L{1.9cm}|L{1.85cm}|Y|L{2.05cm}|Y|}
\hline
\textbf{Work} & \textbf{Setting} & \textbf{Key finding} & \textbf{Evaluation} & \textbf{Gap addressed here} \\ \hline
\citet{maass1992lower} & Half-intervals, boxes on a grid & Half-intervals on $n$ points: exactly $\lfloor\lg n\rfloor$ mistakes & Proofs & Boxes, hence intervals, bounded only up to constants. \\ \hline
\citet{chen1994line} & Boxes in $\{1,\dots,n\}^d$ & Efficient learner with $O(d^2\log n)$ mistakes from per-endpoint searches & Algorithm, proofs & Exact optimum open even for one interval. \\ \hline
\citet{bshouty1998exact} & Unions of $s$ boxes, queries & At most $sd$ counterexamples; equivalence-only learning for constant $d$ & Algorithm, proofs & Optimal count for unions of intervals not determined. \\ \hline
\citet{kimber1995line} & Smooth real functions on $[0,1]$ & Optimal total $p$-th power error exactly $1$ for $p,q\ge2$ & Proofs & Exact values only for real-valued, not binary, classes. \\ \hline
\citet{auer1999structural} & Mistakes with and without queries & Membership queries lower the optimum by at most a factor $1/\log_2(4/3)\approx2.41$; tight & Proofs & Optimal bound of a given class not evaluated. \\ \hline
\citet{alon2022private} & Differentially private PAC learning & A class is privately learnable if and only if its Littlestone dimension is finite & Proofs & Finite versus infinite only; values for concrete classes not computed. \\ \hline
\citet{frances1998optimal} & Explicit finite classes & VC dimension reduces to Littlestone dimension; LOGNP-hard & Reduction & Structured families may still admit closed forms. \\ \hline
\citet{manurangsi2017inapproximability} & Explicit finite classes & Quasi-polynomial exact algorithm; hard to approximate under the randomised Exponential Time Hypothesis & Hardness proofs & No class-specific values. \\ \hline
\citet{filmus2025optimal} & Randomised learners & Optimal expected mistakes equal the randomised Littlestone dimension & Proofs & Not evaluated for thresholds or intervals. \\ \hline
\citet{hanneke2023trichotomy} & Transductive online learning & Mistakes grow as $n$, $\Theta(\log n)$ or $\Theta(1)$ & Proofs & Stated in terms of $d$; $d$ of concrete classes not computed. \\ \hline
\citet{chase2025optimal} & Transductive online learning & Tight $\Omega(\sqrt d)$ transductive bound & Proofs & Takes $d$ of specific classes as given. \\ \hline
\citet{devulapalli2024dimension} & Self-directed learning & Exact self-directed dimension; unions of $j$ intervals cost $2j$ & Proofs, examples & Adversarial order, where cost grows with $m$, not treated. \\ \hline
\citet{herbster1998tracking} & Experts with $k$ shifts & Extra loss $O(k\log n+k\log(\ell/k))$ over best partition & Relative loss bounds & Switches along a fixed domain, realisable case, not analysed. \\ \hline
\citet{attias2026tradeoffs} & Learning with empirical-risk-minimisation oracles & Unions of $j$ intervals on $T$ points: dimension $O(j\log T)$ & Proofs & Dimension of unions of intervals known only to order. \\ \hline
\citet{doron2026theory} & \hspace{0pt}Autoregressive online learning & Intervals and unions named for sharper class-specific bounds & Proofs & Exact base-class dimension not available. \\ \hline
\citet{tassarotti2021formal}; \citet{bagnall2019certifying}; \citet{sonoda2026lean} & Proof assistants (Lean, Coq) & PAC learnability of decision stumps, generalization bounds for trained models, and Rademacher-complexity bounds, machine-checked & Machine-checked proofs & No machine-checked value of a learning dimension. \\ \hline
\citet{domingos2000mining}; \citet{kosina2015very}; \citet{duarte2016adaptive}; \citet{lughofer2022evolving} & Trees and rules learned from data streams & Single-pass learning of threshold splits and rules; Hoeffding trees stay close to the tree a batch learner would build & Experiments; a probabilistic guarantee for Hoeffding trees & No worst-case bound on the mistakes of interval rules. \\ \hline
\textbf{This work} & \textbf{Intervals, unions of $j$ intervals and $k$-switch sequences on $m$ points; deterministic and randomised online learners.} & \textbf{Exact Littlestone dimension for every $k$, with both end labels fixed when $k$ is odd (three-regime law); the natural learner is suboptimal; exact randomised dimension of thresholds, and that of intervals to within $0.64$.} & \textbf{Complete proofs, cross-checked by exhaustive exact computation; the exact formulas are machine-checked in Lean~4.} & \textbf{Addresses the gaps listed above.} \\ \hline
\end{tabularx}
\end{table}

\subsection{The mistake-bound model and its combinatorial characterisation}

The mistake-bound model goes back to \citet{littlestone1988learning}, who showed that the optimal worst-case number of mistakes of a deterministic learner equals the depth of the deepest complete mistake tree shattered by the class, and that the Standard Optimal Algorithm attains it. Finiteness of the same dimension also decides which classes can be PAC learned under approximate differential privacy \citep{alon2022private}. \citet{angluin1988queries} introduced the closely related model of equivalence queries; her majority-vote strategy, like the halving algorithm discussed by Littlestone, makes at most $\lg|H|$ mistakes. We work within this characterisation and add its exact value for intervals, unions of intervals and switch-bounded sequences, together with explicit learners that attain it.

The recursive form of the dimension used in our proofs is standard \citep{ben2009agnostic}, and textbook treatments place it next to the Vapnik--Chervonenkis (VC) dimension \citep{vapnik1971uniform}: the VC dimension never exceeds the Littlestone dimension, and thresholds on $[0,1]$ have VC dimension $1$ but infinite Littlestone dimension \citep{shalev2014understanding}. Intervals have VC dimension $2$, so by the bound of \citet{sauer1972density} the counting bound $\lg|\Int_m|$ is about $2\lg m$. Our exact formula places $\Ldim(\Int_m)$ between $2\lg(m+1)-(2\lg5-2)$ and $2\lg(m+1)-(2\lg7-4)$, strictly below the counting bound for many $m$, with constants that neither the VC dimension nor counting predicts.

\citet{auer1999structural} compared the mistake-bound model with and without membership queries and showed that $\mathrm{opt}_{\mathrm{memb}}(F)\ge\log_2(4/3)\,\mathrm{opt}_{\mathrm{stand}}(F)$ for every class $F$, so membership queries lower the optimal number of mistakes by at most a factor of about $2.41$, and this constant cannot be improved; they also bounded the price of weak reinforcement in the multiclass case. Their results relate models to one another up to constant factors, whereas our concern is the additive constant inside a single model for single classes.

This line of work characterises the optimal mistake bound abstractly but leaves its value for most natural classes undetermined.

\subsection{Online learning of geometric concepts on discrete domains}

\citet{maass1992lower} developed lower-bound methods for online learning, separated several learning models, and proved that the optimal mistake bound for half-intervals on $\{1,\dots,n\}$ is exactly $\lfloor\lg n\rfloor$, so binary search is optimal. For axis-parallel boxes they gave bounds that hold up to constant factors. Our threshold lemma restates their half-interval result for chains of thresholds, and we extend exactness from one boundary to two and to $k$ boundaries, where binary search on each boundary is no longer optimal.

\citet{maass1994algorithms} showed that boxes in $\{1,\dots,n\}^d$ can be learned with $\Theta(\log n)$ mistakes for every fixed $d$ when the hypotheses are boxes, and \citet{chen1992line} established that the optimal mistake bound with arbitrary hypotheses is $\Theta(d\log n)$. In the journal version, \citet{chen1994line} gave an efficient learner with $O(d^2\log n)$ mistakes that, after the first positive counterexample, runs $2d$ error-tolerant binary searches, one for each endpoint parameter of the target box, and an asymptotically optimal learner for unions of two boxes in the plane. For $d=1$ our results replace $\Theta(\log n)$ by an exact formula, and they show that a learner of this shape (wait for a positive example, then search for each endpoint separately) is not exactly optimal for intervals.

\citet{bshouty1998exact} learned unions of $s$ discretised boxes in $d$ dimensions from membership and equivalence queries with at most $sd$ counterexamples, and from equivalence queries alone in polynomial time for constant dimension. Their bounds are upper bounds on the number of counterexamples of particular efficient algorithms; for $d=1$ we determine the optimal number of mistakes for unions of $j$ intervals, which equals $2j\lg\frac{m+1}{j}$ up to an additive $O(j)$ term.

These studies settle the order of growth for boxes and unions of boxes, but not the exact optimum even on a line.

\subsection{Exact and sharp mistake bounds for concrete classes}

Exact optimal values are known in a few related models. \citet{kimber1995line} studied online prediction of real-valued functions on $[0,1]$ whose derivative has bounded $q$-norm and showed that the optimal worst-case sum of $p$-th powers of the errors is exactly $1$ for all $p,q\ge2$, but infinite for $p=1$. \citet{long2000improved} determined the trial-dependent version for $q=2$ and absolute errors to within an additive constant, and \citet{geneson2023online} determined the optimum up to constant factors in new ranges of $(p,q)$ and found the exact value $1$ for $q\in(1,2)$ and $p\ge2+\frac{1}{q-1}$. Our results are of the same exact kind but for binary classes on a discrete domain, where the optimum is an integer-valued function of $m$ with oscillating constants rather than a constant.

Another strand compares feedback models. In apple tasting \citep{helmbold2000apple} the learner sees the label only when it predicts a positive, and \citet{geneson2021note} repaired a lemma in Long's proof that, for infinitely many label-set sizes $|Y|$, the bandit mistake bound in multiclass online learning can exceed the standard one by a factor $(1-o(1))|Y|\ln|Y|$. These results compare optimal values across feedback models, while ours computes the optimal value itself for a family of classes under full feedback.

Outside smooth real-valued functions and threshold-like chains, exact optimal mistake bounds for natural classes remain scarce.

\subsection{Computing the Littlestone dimension}

\citet{frances1998optimal} reduced the computation of the VC dimension to that of the Littlestone dimension, so that the latter is hard for the class LOGNP when the class is given explicitly. \citet{manurangsi2017inapproximability} noted that both dimensions can be computed exactly in quasi-polynomial time and proved, under the randomised Exponential Time Hypothesis, nearly matching lower bounds that hold even for approximation algorithms. These results explain why our computational checks use structural compressions (segment states, gap multisets, a cutting game) rather than the generic recursion beyond $m\le16$.

\citet{kozachinskiy2024simple} studied learners that access the class only through a consistent oracle, which avoids computing the dimension, and gave a learner with $O(256^d)$ mistakes together with a lower bound of $3^d$ in that model. Our results go the other way for specific classes: they compute the dimension exactly, and the matching upper bounds come from an explicit block-pairing learner rather than from an oracle.

Hardness results concern general explicitly given classes; they say nothing about closed forms for structured families such as intervals.

\subsection{Randomised, transductive and self-directed variants}

\citet{filmus2025optimal} proved that the optimal expected mistake bound of a randomised learner equals the randomised Littlestone dimension, defined through the average depth of shattered trees, and determined the optimal randomised agnostic bound $k+\Theta(\sqrt{kd}+d)$; its deterministic consequence $2k+\Theta(d)+O(\sqrt{kd})$ resolved a question studied by \citet{auer1999structural}. We use their characterisation and recursion as given, compute the randomised dimension of thresholds exactly, and prove an explicit lower bound for intervals that matches exact computation for $m\le70$, with an upper bound less than $0.64$ above it.

In the transductive model of \citet{ben1997online}, the learner knows the sequence of instances in advance. \citet{hanneke2023trichotomy} proved that the optimal number of mistakes then grows as $n$, $\Theta(\log n)$ or $\Theta(1)$, decided by the VC and Littlestone dimensions, and improved the lower bound in the constant case to $\Omega(\log d)$. \citet{chase2025optimal} closed the question with a tight $\Omega(\sqrt d)$ bound and a matching construction. Both results are stated in terms of the Littlestone dimension $d$ of a general class; our formulas give $d$ exactly for intervals and their unions.

When the learner chooses the order of instances, geometric classes become far cheaper. \citet{goldman1994power} showed that axis-parallel boxes can be learned with two mistakes, and \citet{devulapalli2024dimension} introduced a dimension that characterises the self-directed mistake bound exactly and computed the value $2j$ for unions of $j$ intervals (their $k$-interval classifiers). In the adversarial order studied here, unions of $j$ intervals on $m$ points cost about $2j\lg(m/j)$ mistakes, and our results pin down the additive constants.

Exact values of the randomised Littlestone dimension for concrete geometric classes such as thresholds and intervals have not been computed in this line of work.

\subsection{Intervals, unions of intervals and switching in recent work}

\citet{attias2026tradeoffs} studied the tradeoff between mistakes and calls to empirical-risk-minimisation or weak-consistency oracles in online and transductive learning; for unions of $j$ intervals on $T$ points (their $k$-intervals) they bound the optimal number of mistakes by $O(j\log T)$. \citet{doron2026theory} developed an online theory of autoregressive generation and named intervals and unions of intervals among the concrete classes for which sharper, class-specific mistake bounds may be expected. Our results give the exact Littlestone dimension of these base classes in the standard model, which the first states only up to order and the second leaves open; they do not address the oracle or autoregressive settings themselves.

Sequences with few label switches are the concept-class counterpart of shifting targets. \citet{herbster1998tracking} bounded the additional loss over the best partition of a sequence into $k+1$ segments by $O(k\log n+k\log(\ell/k))$ for $n$ experts. In their setting the switches occur in time and the bounds are relative and asymptotic; in ours the switches occur along a fixed domain whose points the adversary presents in any order, and the realisable optimum is determined exactly for every even $k$ and for $k=1$.

Recent work uses intervals and unions of intervals as running examples, but records their dimension only as $O(j\log T)$.

\subsection{Machine-checked learning theory}

Proof assistants have been used for a few results of statistical learning theory. \citet{tassarotti2021formal} proved in Lean that decision stumps are PAC learnable, \citet{bagnall2019certifying} built a system in Coq that proves generalization bounds for trained models, and \citet{sonoda2026lean} formalised generalization bounds through Rademacher complexity and Dudley's entropy integral in Lean~4. These works verify sample-complexity guarantees. We are not aware of a machine-checked exact value of a learning dimension for a natural class; Section~\ref{sec:verification} gives such values for intervals, unions of intervals and switch-bounded sequences.

\subsection{Research gaps and contributions}

The work summarised in Table~\ref{tab:comparison} leaves three gaps. For the simplest geometric classes beyond thresholds (intervals, unions of intervals and sequences with few switches) only the order of growth of the optimal mistake bound, $\Theta(\log m)$ or $O(k\log m)$, is known, and no exact value \citep{maass1992lower,chen1994line,attias2026tradeoffs}. Exact optimal mistake bounds are also known for very few natural classes \citep{kimber1995line,geneson2023online}, and computing the dimension is hard in general \citep{frances1998optimal,manurangsi2017inapproximability}, so it has been unclear whether natural learners, such as binary search on each boundary, are exactly optimal. Finally, the randomised Littlestone dimension is characterised abstractly \citep{filmus2025optimal}, but exact values for concrete geometric classes such as thresholds and intervals have not been computed.

We address the first gap by determining the Littlestone dimension exactly for intervals, unions of $j$ intervals and binary sequences with at most $k$ switches on a line (for every even $k$, and for every $k$ when both end labels are fixed), through a three-regime law for the smallest domain that forces $d$ mistakes; the proof combines a gap potential and a block-pairing learner for the upper bound with explicit adversary strategies and a refinement argument for the lower bound. The natural learner, which waits for a positive example and then binary-searches both endpoints, turns out not to be optimal, and we quantify by how much; this settles the second gap. For the third, we determine the randomised Littlestone dimension of thresholds and, for intervals, prove a greedy-adversary lower bound and an upper bound that differ by less than $0.64$; we conjecture that the lower bound is exact. Every statement is cross-checked by exhaustive exact computation, the exact formulas are also machine-checked in Lean~4, and the last row of Table~\ref{tab:comparison} summarises these contributions against the prior work.

\section{Preliminaries}\label{sec:prelim}

For a class $H$ of functions $X\to\{0,1\}$, a point $x\in X$ and $b\in\{0,1\}$, write $H_x^b=\{h\in H:h(x)=b\}$. We assume that the adversary never presents a point twice: a learner that remembers the labels it has seen makes no mistake on a repeated point, so repetitions change no mistake bound. We use the standard recursive characterisation of the Littlestone dimension (e.g.\ \cite{littlestone1988learning,ben2009agnostic,shalev2014understanding}).

\begin{fact}\label{fact:rec}
Let $H\neq\emptyset$.
\begin{enumerate}[label=(\roman*)]
\item $\Ldim(H)\ge d+1$ if and only if there is a point $x\in X$ with $H_x^0,H_x^1\neq\emptyset$ and
\[
\Ldim(H_x^0)\ge d,\qquad \Ldim(H_x^1)\ge d.
\]
\item $\Ldim(H)=0$ when $|H|=1$.
\item $\Ldim$ is monotone under inclusion. It is invariant under bijective relabelling of the domain and under deleting domain points on which all of $H$ agrees.
\end{enumerate}
\end{fact}

\begin{lemma}[Thresholds]\label{lem:thr}
For $c\ge 1$ let
\[
\mathrm{Th}_c=\bigl\{\,y\mapsto \mathbf 1[y\ge a]:\ a\in[c]\,\bigr\}
\]
on the domain $[c-1]$. Then $\Ldim(\mathrm{Th}_c)=\lgf{c}$. The same holds for the mirrored class $\{y\mapsto\mathbf 1[y\le a]\}$.
\end{lemma}
\begin{proof}
Every version space reachable from $\mathrm{Th}_c$ has the form $\{h_a:a\in R\}$ for a set $R$ of $s$ consecutive integers. Querying a point splits $R$ into a nonempty prefix and a nonempty suffix of sizes $s_1+s_2=s$, and every such split is available. Writing $L(s)$ for the resulting dimension,
\[
L(1)=0,\qquad L(s)=1+\max_{1\le s_1<s}\min\bigl(L(s_1),\,L(s-s_1)\bigr).
\]
By induction $L$ is nondecreasing and $L(s)=\lgf{s}$. The split $s_1=\lfloor s/2\rfloor$ gives $1+\lgf{\lfloor s/2\rfloor}=\lgf{s}$, and no split does better, because $\min(s_1,s-s_1)\le\lfloor s/2\rfloor$.
\end{proof}

\begin{lemma}[Products]\label{lem:prod}
Let $H_1,H_2$ be nonempty classes on disjoint domains $X_1,X_2$, and let
\[
\begin{aligned}
H_1\otimes H_2=\bigl\{h:X_1\cup X_2\to\{0,1\}:\ &h|_{X_1}\in H_1,\\ &h|_{X_2}\in H_2\bigr\}.
\end{aligned}
\]
Then
\[
\Ldim(H_1\otimes H_2)=\Ldim(H_1)+\Ldim(H_2).
\]
\end{lemma}
\begin{proof}
\emph{Lower bound.} Graft a shattered tree for $H_2$ below every leaf of a shattered tree for $H_1$.

\emph{Upper bound.} We use induction on $|H_1|+|H_2|$ (all classes here are finite). Suppose $\Ldim(H_1\otimes H_2)\ge d+1$. Fact~\ref{fact:rec} gives a point $x$, say $x\in X_1$, with
\[
\Ldim\bigl((H_1)^b_x\otimes H_2\bigr)\ge d\qquad\text{for both } b\in\{0,1\}.
\]
By induction $\Ldim((H_1)^b_x)\ge d-\Ldim(H_2)$ for both $b$, and Fact~\ref{fact:rec} gives $\Ldim(H_1)\ge d+1-\Ldim(H_2)$.
\end{proof}

\section{From two switches to intervals}\label{sec:switch}

\begin{lemma}\label{lem:cyc}
Let $n\ge 2$, $j\ge 1$ and $p\in[n]$, and list $[n]\setminus\{p\}$ in cyclic order after $p$ as
\[
\sigma=(p+1,\dots,n,1,\dots,p-1).
\]
Then $f\mapsto f\circ\sigma$ is a bijection
\[
\{f\in C_{n,2j}:\ f(p)=0\}\ \longrightarrow\ U_j(n-1).
\]
In particular $R_{2j}(n)\cong U_j(n-1)$.
\end{lemma}
\begin{proof}
For $f:[n]\to\{0,1\}$ let $s(f)$ be its number of switches along the path $1,2,\dots,n$, and let
\[
c(f)=s(f)+\mathbf 1[f(n)\ne f(1)]
\]
be its number of switches around the cycle $1,\dots,n,1$. Then $c(f)$ is even and $s(f)\le c(f)\le s(f)+1$, hence
\[
s(f)\le 2j\iff c(f)\le 2j.
\]
Now read the cycle starting and ending at $p$, where $f(p)=0$. Then $c(f)$ is the number of switches of the sequence
\[
0,\ f(\sigma_1),\ \dots,\ f(\sigma_{n-1}),\ 0,
\]
which is twice the number of maximal runs of ones in $f\circ\sigma$. So $f\in C_{n,2j}$ if and only if $f\circ\sigma$ has at most $j$ runs of ones, that is, $f\circ\sigma\in U_j(n-1)$. Finally, the map is a bijection between $\{f:f(p)=0\}$ and $\{0,1\}^{n-1}$. The functions with $f(p)=0$ and at most $2j$ switches around the cycle form $R_{2j}(n)$ with fixed point $p$, which gives the last claim.
\end{proof}

\begin{proof}[Proof of Theorem~\ref{thm:switch}]
For $n=1$ both sides equal $1$. Let $n\ge2$ and $p\in[n]$.

Complementation $f\mapsto 1-f$ preserves $C_{n,2j}$ and maps $(C_{n,2j})^0_p$ onto $(C_{n,2j})^1_p$, so both have the same dimension. By Lemma~\ref{lem:cyc} and Fact~\ref{fact:rec} (relabelling),
\[
\Ldim\bigl((C_{n,2j})^0_p\bigr)=\Ldim(U_j(n-1))\qquad\text{for every } p.
\]
Fact~\ref{fact:rec} then gives $\Ldim(C_{n,2j})=1+\Ldim(U_j(n-1))$. The second statement follows from Theorem~\ref{thm:main} with ${m=n-1}$.
\end{proof}

\section{From intervals to a cutting game}\label{sec:cut}

\begin{definition}[Segment classes]
For a finite multiset $S$ of positive integers, place disjoint blocks $B_g$ of consecutive points, one of length $g$ for each $g\in S$. Let $J_S$ consist of $\emptyset$ together with every set of consecutive points contained in a single block. Up to relabelling, $J_S$ depends only on $S$, and $\Int_m=J_{\{m\}}$.
\end{definition}

\begin{lemma}\label{lem:branch}
Let $x$ be the $i$-th point of a block of length $g$ in $S$. Then:
\begin{enumerate}[label=(\alph*)]
\item $\Ldim\bigl((J_S)^1_x\bigr)=\lgf{i}+\lgf{g-i+1}$;
\item $(J_S)^0_x=J_{S'}$, where $S'=S-\{g\}+\{i-1,\,g-i\}$, with zeros discarded.
\end{enumerate}
\end{lemma}
\begin{proof}
(a) The members of $J_S$ containing $x$ are the sets $\{a,\dots,b\}$ inside that block with $a\le x\le b$, and they vanish outside the block.
\begin{itemize}[leftmargin=*]
\item On the $i-1$ points to the left of $x$ they are the thresholds $\mathbf 1[y\ge a]$, with $i$ choices of $a$.
\item On the $g-i$ points to the right of $x$ they are the mirrored thresholds, with $g-i+1$ choices of $b$.
\end{itemize}
The two choices are independent, so Lemmas~\ref{lem:thr} and~\ref{lem:prod} give $\lgf{i}+\lgf{g-i+1}$. The same argument shows more generally that the intervals $[a,b]$ with $a_1\le a\le a_2\le b_1\le b\le b_2$ form a class of dimension $\lgf{a_2-a_1+1}+\lgf{b_2-b_1+1}$.

(b) The members avoiding $x$ are $\emptyset$ and the sets inside some block not crossing $x$. On the remaining domain this is $J_{S'}$.
\end{proof}

\begin{figure}[!htbp]
\centering
\includegraphics[width=\textwidth]{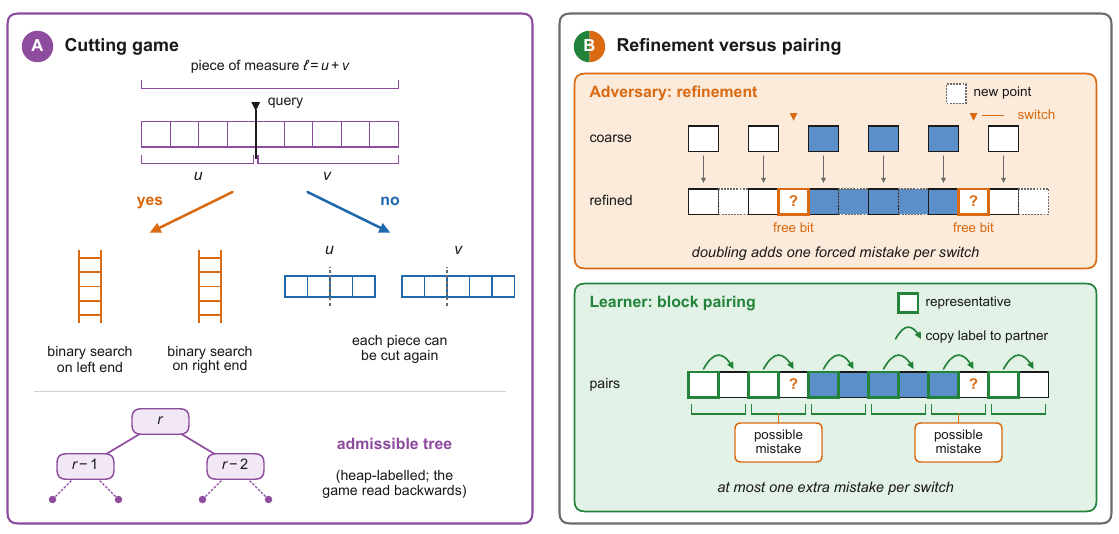}
\caption{The two mechanisms behind the bounds. (A) The cutting game. A query cuts a piece of measure $\ell=u+v$. The label $1$ (``yes'') leaves two independent binary searches, one per endpoint; the label $0$ (``no'') leaves the pieces $u$ and $v$, which can be cut again. Read backwards, an adversary strategy is a heap-labelled admissible tree. (B) Refinement and pairing. The adversary inserts a new point in every slot; a new point in a slot that holds a switch can take either label without adding a switch, which forces one more mistake per switch (Lemma~\ref{lem:moves}(e)). The learner pairs neighbouring points, learns on one representative per pair and copies its label to the partner. A copy can be wrong only in a pair that contains a switch, so doubling the domain costs the learner at most one extra mistake per switch (Lemma~\ref{lem:pair}).}
\label{fig:mechanism}
\end{figure}

Give each block of length $g$ the \emph{measure} $g+1$. Querying the $i$-th point of a block of measure $\ell=g+1$ then corresponds to a \emph{cut} $\ell=u+v$ with $u=i\ge1$, ${v=g-i+1\ge1}$; by Lemma~\ref{lem:branch} the positive branch has dimension $\lgf u+\lgf v$, which we call the \emph{value} of the cut, and the negative branch replaces the piece $\ell$ by pieces of measures $u$ and $v$. Pieces of measure $1$ (empty blocks) cannot be cut. Figure~\ref{fig:mechanism}(A) shows one cut and its two branches.

\begin{lemma}\label{lem:cut}
$\Ldim(J_S)\ge d$ if and only if, starting from the pieces $\{g+1:g\in S\}$, there is a sequence of $d$ cuts, each applied to a current piece, such that
\[
\text{the $i$-th cut has value at least } d-i\qquad(i=1,\dots,d).
\]
\end{lemma}
\begin{proof}
Induction on $d$; $d=0$ is trivial. By Fact~\ref{fact:rec} and Lemma~\ref{lem:branch}, $\Ldim(J_S)\ge d$ if and only if some query $x$ satisfies
\[
\Ldim\bigl((J_S)^1_x\bigr)\ge d-1\quad\text{and}\quad \Ldim(J_{S'})\ge d-1 .
\]
The first condition says that the cut has value at least $d-1$. By induction, the second holds if and only if $S'$ admits $d-1$ further cuts with values at least $d-2,\dots,0$.
\end{proof}

\begin{definition}[Admissible trees]
An \emph{admissible tree} is a finite full binary tree with positive integer measures $\mu$ on its nodes such that:
\begin{enumerate}[label=(\roman*)]
\item the internal nodes carry distinct labels in $\N$, and each internal child has a smaller label than its parent (heap order);
\item $\mu(v)=\mu(v_1)+\mu(v_2)$ for every internal node $v$ with children $v_1,v_2$;
\item every internal node $v$ with children $v_1,v_2$ satisfies
\[
\lgf{\mu(v_1)}+\lgf{\mu(v_2)}\;\ge\;\mathrm{label}(v).
\]
\end{enumerate}
Let $P^\ast(d)$ be the minimum root measure of an admissible tree whose labels are exactly $\{0,\dots,d-1\}$; thus $P^\ast(0)=1$, a single leaf.
\end{definition}

\begin{lemma}\label{lem:tree}
$\Ldim(\Int_m)\ge d$ if and only if $P^\ast(d)\le m+1$.
\end{lemma}
\begin{proof}
($\Rightarrow$) Take the $d$ cuts of Lemma~\ref{lem:cut} on the single rod of measure $m+1$. Make every piece a node, make the two pieces produced by a cut the children of the piece that was cut, and give the $i$-th cut the label $d-i$. A piece is created before it is cut, so labels decrease downwards, and the value condition is the admissibility inequality.

($\Leftarrow$) Increase one leaf until the root has measure exactly $m+1$; since $\lambda$ is nondecreasing, admissibility is preserved. Perform the cuts in decreasing order of labels. The parent of a node has a larger label, so it has already been cut when the node's turn comes, and the $i$-th cut performed has label $d-i$ and value at least $d-i$. Lemma~\ref{lem:cut} finishes the proof.
\end{proof}

\section{Solving the tree problem}\label{sec:tree}

\begin{lemma}[Elementary bounds]\label{lem:elem}
\begin{enumerate}[label=(\alph*)]
\item For $r\ge 0$ let $c(r):=\min\{u+v:\ \lgf u+\lgf v\ge r\}$. Then
\[
c(r)=2^{\lfloor r/2\rfloor}+2^{\lceil r/2\rceil},
\]
and every internal node with label $r$ has measure at least $c(r)$.
\item For $a,b\ge1$ and $r\ge0$ let
\[
g(a,b,r):=\min\{u+v:\ u\ge a,\ v\ge b,\ \lgf u+\lgf v\ge r\}.
\]
Then
\[
g(a,b,r)=\min_{0\le p\le r}\bigl[\max(a,2^p)+\max(b,2^{r-p})\bigr],
\]
and $g$ is nondecreasing in $a$ and in $b$.
\item (Relabelling) If an admissible tree has $s$ internal nodes, replacing the labels by their ranks $0,\dots,s-1$ keeps it admissible. Hence every admissible tree (or subtree) with $s$ internal nodes has root measure at least $P^\ast(s)$.
\end{enumerate}
\end{lemma}
\begin{proof}
(b) The choice $u=\max(a,2^p)$, $v=\max(b,2^{r-p})$ is feasible. Conversely, for a feasible pair $(u,v)$ let $p=\min(\lgf u,r)$. Then $\lgf v\ge r-p$, so
\[
u\ge\max(a,2^p)\quad\text{and}\quad v\ge\max(b,2^{r-p}).
\]
Monotonicity in $a$ and $b$ follows from this formula.

(a) This is (b) with $a=b=1$, since $2^p+2^{r-p}$ is minimised at $p=\lfloor r/2\rfloor$.

(c) Ranking preserves the order of labels and does not increase any label.
\end{proof}

\begin{proposition}[Upper bound]\label{prop:upper}
$P^\ast(d)\le P(d)$ for all $d$. More precisely, for every $d\ge3$ there is an admissible tree with
\begin{itemize}[leftmargin=*]
\item labels $\{0,\dots,d-1\}$,
\item root measure $P(d)$, and
\item a leaf of measure at least $2$.
\end{itemize}
\end{proposition}
\begin{proof}
Write a node as $[r]\langle\cdot,\cdot\rangle$ with its label $r$ and its two subtrees, and write leaves as their measures.

\emph{Base trees.} Let $T_0$ be a single leaf of measure $1$, and
\[
\begin{aligned}
T_1&=[0]\langle1,1\rangle,\\
T_2&=[1]\langle T_1,1\rangle,\\
T_3&=[2]\bigl\langle [0]\langle1,1\rangle,\ [1]\langle1,2\rangle\bigr\rangle,\\
T_4&=[3]\bigl\langle [2]\langle [0]\langle1,1\rangle,2\rangle,\ [1]\langle1,2\rangle\bigr\rangle,
\end{aligned}
\]
with root measures $2$, $3$, $5$ and $7$. The conditions to check are
\[
\begin{aligned}
T_3&:\quad \lgf2+\lgf3=2,\qquad \lgf1+\lgf2=1,\\
T_4&:\quad \lgf4+\lgf3=3,\qquad \lgf2+\lgf2=2,
\end{aligned}
\]
every node $[0]\langle1,1\rangle$ is a copy of $T_1$, and the node $[1]\langle1,2\rangle$ of $T_4$ is as in $T_3$. Both $T_3$ and $T_4$ have a leaf of measure $2$.

\emph{Doubling step.} Let $T$ be admissible with labels $\{0,\dots,d-3\}$, root measure $M$ and a leaf of measure $w\ge2$. Multiply every measure by $2$ and add $2$ to every label. Since $\lgf{2x}=\lgf x+1$, every inequality gains $2$ on both sides. Now replace the leaf of measure $2w\ge4$ by the subtree
\[
[1]\bigl\langle [0]\langle1,1\rangle,\ 2w-2\bigr\rangle .
\]
Its new node of label $1$ satisfies $\lgf2+\lgf{2w-2}\ge 1+1\ge 1$, the new nodes lie below a node of label at least $2$, and the measure is unchanged. The result is admissible, with labels $\{0,\dots,d-1\}$, root measure $2M$ and a leaf of measure $2w-2\ge2$.

\emph{Conclusion.} Start from $T_3$ and $T_4$ and apply the doubling step repeatedly. Since $P(d)=2P(d-2)$ for $d\ge5$, this gives, for every $d\ge3$, an admissible tree with labels $\{0,\dots,d-1\}$, root measure $P(d)$ and a leaf of measure at least $2$, along the sequence $5,7,10,14,\dots$. Hence $P^\ast(d)\le P(d)$ for $d\ge 3$. The cases $d\le 2$ are $T_0$, $T_1$ and $T_2$.
\end{proof}

\begin{table}[!htbp]
\centering
\caption{Case analysis in the proof of Proposition~\ref{prop:lower}, for $t\ge2$. Each row bounds $\max(a,2^p)+\max(b,2^q)$, with $p+q$ equal to the third argument of $g$, on one range of $p$. The bounds and the values of $P(d-1)$, $c(d-2)$, $c(d-3)$ and $a_0$ are in units of $2^{t-2}$; the literal arguments $1$ and $2$ of $g$ and the parameters $r$, $p$, $q$ are not. The target $P(d)$ is $10$ units for $d=2t+1$ and $7$ units for $d=2t$.}
\label{tab:lowercases}
\begin{tabular}{|c|c|c|c|}
\hline
\textbf{Case} & \textbf{Quantity} & \textbf{Range} & \textbf{Bound} \\ \hline
\multicolumn{4}{|l|}{Odd $d=2t+1$: $r=2t$, $P(d-1)=7$, $c(d-2)=6$, $c(d-3)=4$} \\ \hline
\multirow{2}{*}{(i)} & \multirow{2}{*}{$g(P(d-1),1,r)$} & $p\le t$ & $11$ \\ \cline{3-4}
 & & $p\ge t+1$ & $10$ \\ \hline
\multirow{2}{*}{(ii)} & \multirow{2}{*}{$g(c(d-2),c(d-3),r)$} & $p\le t$ & $10$ \\ \cline{3-4}
 & & $p\ge t+1$ & $12$ \\ \hline
\multirow{2}{*}{(iii)} & \multirow{2}{*}{$a_0=g(c(d-3),1,r-1)$} & $p\le t$ & $6$ \\ \cline{3-4}
 & & $p\ge t+1$ & $8$ \\ \hline
\multirow{3}{*}{(iii)} & \multirow{3}{*}{$g(a_0,2,r)$, $a_0\ge6$} & $p\le t$ & $10$ \\ \cline{3-4}
 & & $p=t+1$ & $10$ \\ \cline{3-4}
 & & $p\ge t+2$ & $16$ \\ \hline
\multicolumn{4}{|l|}{Even $d=2t$: $r=2t-1$, $P(d-1)=5$, $c(d-2)=4$, $c(d-3)=3$} \\ \hline
\multirow{2}{*}{(i)} & \multirow{2}{*}{$g(P(d-1),1,r)$} & $p\le t$ & $7$ \\ \cline{3-4}
 & & $p\ge t+1$ & $8$ \\ \hline
\multirow{2}{*}{(ii)} & \multirow{2}{*}{$g(c(d-2),c(d-3),r)$} & $p\le t$ & $7$ \\ \cline{3-4}
 & & $p\ge t+1$ & $11$ \\ \hline
\multirow{3}{*}{(iii)} & \multirow{3}{*}{$a_0=g(c(d-3),1,r-1)$} & $p\le t-1$ & $5$ \\ \cline{3-4}
 & & $p=t$ & $5$ \\ \cline{3-4}
 & & $p\ge t+1$ & $8$ \\ \hline
\multirow{2}{*}{(iii)} & \multirow{2}{*}{$g(a_0,2,r)$, $a_0\ge5$} & $p\le t$ & $7$ \\ \cline{3-4}
 & & $p\ge t+1$ & $8$ \\ \hline
\end{tabular}
\end{table}

\begin{proposition}[Lower bound]\label{prop:lower}
$P^\ast(d)\ge P(d)$ for all $d$.
\end{proposition}
\begin{proof}
\emph{Small $d$.} By Lemma~\ref{lem:elem}(a), $P^\ast(1)\ge c(0)=2$ and $P^\ast(2)\ge c(1)=3$. For $d=3$, the node labelled $1$ must be a child of the root, because only label $2$ is larger; so it has measure $\mu_1\ge c(1)=3$. The other child has measure $\mu_2\ge1$, and $\lgf{\mu_1}+\lgf{\mu_2}\ge2$. Hence either $\mu_1=3$ and $\mu_2\ge2$, or $\mu_1\ge4$; in both cases the root has measure at least $5$.

\emph{Three cases.} Now let $d\ge4$ and assume the claim for $d-1$. Let $\rho$ be the root (label $d-1$) of an admissible tree with labels $\{0,\dots,d-1\}$. The node labelled $d-2$ has a parent with a larger label, so it is a child $X$ of $\rho$; let $Y$ be the other child. The node labelled $d-3$ has parent $\rho$ or $X$, which leaves three cases.
\begin{enumerate}[label=(\roman*)]
\item $Y$ is a leaf. Then $X$ carries all $d-1$ other labels, so $\mu(X)\ge P^\ast(d-1)\ge P(d-1)$ by Lemma~\ref{lem:elem}(c) and induction, and
\[
\mu(\rho)\ge g\bigl(P(d-1),1,d-1\bigr).
\]
\item $Y$ is the node labelled $d-3$. Then
\[
\mu(\rho)\ge g\bigl(c(d-2),c(d-3),d-1\bigr).
\]
\item $Y$ is internal with label below $d-3$. Then the node labelled $d-3$ is a child of $X$, so $\mu(X)\ge a_0:=g(c(d-3),1,d-2)$, while $\mu(Y)\ge c(0)=2$. Hence
\[
\mu(\rho)\ge g(a_0,2,d-1).
\]
\end{enumerate}

\emph{Evaluating the bounds.} Each of the three bounds is at least $P(d)$. By Lemma~\ref{lem:elem}(b), each is a minimum over splits $r=p+q$ of $\max(a,2^p)+\max(b,2^q)$, so it suffices to bound this sum on a few ranges of $p$. Table~\ref{tab:lowercases} gives these bounds for $t\ge2$ in units of $2^{t-2}$. Each entry follows by inserting the smallest values of $2^p$ and $2^q$ allowed in its range, with one exception. For odd $d$ in case (i) with $p\ge t+1$, the sum is at least
\[
2^{2t-q}+2^q\ \ge\ 2^{t+1}+2^{t-1},
\]
because $q\mapsto 2^{2t-q}+2^q$ decreases for $q<t$. In case (iii) the table bounds $a_0$ first and then uses that $g$ is nondecreasing in $a$. Every entry is at least the target $P(d)$, so $\mu(\rho)\ge P(d)$.
\end{proof}

\paragraph{Three cuts suffice.} The induction in the proof of Proposition~\ref{prop:lower} can be avoided. A rod that admits three successive cuts of values at least $d-1$, $d-2$ and $d-3$ already has measure at least $P(d)$, because $P(d)=c(d-2)+c(d-3)$ for $d\ge3$ (for example $5=3+2$ and $7=4+3$). If the third cut is made in the piece that was not cut second, the two pieces of the first cut have measures at least $c(d-2)$ and $c(d-3)$. Otherwise the third cut is inside a part $a$ of the piece cut second, so $a\ge c(d-3)$, and a case analysis on the values $\lgf{\cdot}$ of the first two cuts gives the same bound. The formal proof of Section~\ref{sec:verification} follows this route.

\begin{proof}[Proof of Theorem~\ref{thm:main}]
Propositions~\ref{prop:upper} and~\ref{prop:lower} give $P^\ast=P$, and Lemma~\ref{lem:tree} gives the first formula.

For the closed form let $\ell=m+1\ge4$ and note that $P$ is increasing. Then
\[
\begin{aligned}
P(2t)\le\ell&\iff t\le\lg\frac{4\ell}{7}\qquad(t\ge1),\\
P(2t+1)\le\ell&\iff t\le\lg\frac{2\ell}{5}\qquad(t\ge1).
\end{aligned}
\]
For $t\ge2$ the first line follows from $P(2t)=7\cdot2^{t-2}$. For $t=1$ both sides of the first line hold, since $P(2)=3\le\ell$ and $\lg\frac{4\ell}{7}\ge 1$ when $\ell\ge4$. Since $P(2)\le\ell$, the maximum $d$ is at least $2$, and it is the larger of $A_m$ and $B_m$ in Theorem~\ref{thm:main}.
\end{proof}

\begin{proof}[Proof of Corollary~\ref{cor:asym}]
Let $\ell=m+1\ge4$ and $d=\Ldim(\Int_m)\ge2$, so $P(d)\le\ell<P(d+1)$. For $e\ge3$,
\[
2\lg P(e)-e=\begin{cases}2\lg7-4\approx1.615, & e\ \text{even},\\ 2\lg5-3\approx1.644, & e\ \text{odd}.\end{cases}
\]

\emph{Upper bound.} We have $2\lg\ell-d\ge2\lg P(d)-d$, which is at least $2\lg7-4$ if $d\ge3$. If $d=2$ then $m=3$ and $2\lg 4-2=2$.

\emph{Lower bound.} Since $d+1\ge3$,
\[
\begin{aligned}
2\lg\ell-d&<2\lg P(d+1)-(d+1)+1\\
&\le(2\lg5-3)+1=2\lg5-2.
\end{aligned}
\]
Equality in the upper bound holds at $\ell=P(2t)$, $t\ge2$.
\end{proof}

\section{Arbitrary switch budgets}\label{sec:general}

Throughout this section $k\ge1$, and $F_k(x)$ denotes $R_k(x)$ when $k$ is even and $G_k(x)$ when $k$ is odd; $L_k(x)=\Ldim(F_k(x))$. We also set $F_0(x)=R_0(x)$, which contains only the constant $0$, so $L_0\equiv0$. In $F_k(x)$ the fixed points are the point labelled $0$ (and, for odd $k$, the point labelled $1$); a \emph{slot} is a pair of consecutive points, and $F_k(x)$ has exactly $x$ slots. The proofs below apply without change to sequences $f(0),\dots,f(x)$ with any fixed end labels $f(0)=a$ and $f(x)=b$ and at most $k$ switches, provided $k\equiv\mathbf 1[a\ne b]\pmod 2$; complementing all labels turns such a class into $R_k(x)$ or $G_k(x)$, so its dimension is also $v_k(x)$.

\subsection{Gap states and a potential}

Consider points on a cycle or on a path, some of them labelled, where the target can be any completion with at most $k$ switches in total (around the cycle, or along the path). We always assume that at least one point is labelled.

A \emph{gap} is the (possibly empty) run of unlabelled points strictly between two consecutive labelled points, taken cyclically on a cycle, or the run between a labelled point and a free end of a path. Empty gaps between adjacent labelled points are counted. There are three types.
\begin{itemize}[leftmargin=*]
\item Type $A$: the two neighbours are labelled points with different labels. A completion places an odd number of switches in such a gap (counting the two boundary slots); an empty $A$ gap holds exactly one forced switch.
\item Type $B$: the two neighbours carry equal labels. A completion places an even number of switches here. On a cycle with a single labelled point, its one gap is of type $B$, bounded by that point on both sides.
\item Type $E$: one neighbour is a labelled point and the other side is a free end of the path. A completion may place any number of switches here.
\end{itemize}
Let $r=k-\#A$ be the \emph{spare budget}, the number of switches still available beyond the one forced in each $A$ gap (empty or not); for a nonempty version space $r\ge0$. The version space is determined by $r$ and the multisets of gap lengths.

\begin{lemma}[Potential]\label{lem:pot}
For such a nonempty version space $V$ (with at least one labelled point) let
\[
\Phi(V)=\Bigl\lfloor\frac{r+\sum_E e}{2}\Bigr\rfloor+\sum_A\Bigl\lceil\frac a2\Bigr\rceil+\sum_B\Bigl\lfloor\frac b2\Bigr\rfloor,
\]
the sums running over the lengths of the gaps of each type. Then $\Phi(V)\ge0$ and $\Ldim(V)\le\Phi(V)$.
\end{lemma}
\begin{proof}
\emph{Reduction to one query.} We show that for every point $x$ for which both $V^0_x$ and $V^1_x$ are nonempty,
\[
\min_{b\in\{0,1\}}\Phi(V^b_x)\le\Phi(V)-1 .
\]
The claim then follows by induction on the number of unlabelled points. If $\Ldim(V)=D\ge1$, Fact~\ref{fact:rec} gives such an $x$ with $\Ldim(V^b_x)\ge D-1$ for both $b$, and for the branch with the smaller potential, $D-1\le\Phi(V^b_x)\le\Phi(V)-1$.

Let $x$ split its gap into $l$ points on one side and $l'$ on the other. There are three cases, by the type of the gap.

\emph{Case 1: $x$ in a $B$ gap of length $b$, so $l+l'=b-1$.} The equal label gives two $B$ gaps and the same $r$. The other label gives two $A$ gaps and spare budget $r-2$, which lowers the first term of $\Phi$ by exactly $1$. The gap terms change as follows.
\begin{itemize}[leftmargin=*]
\item $l,l'$ both odd: the first branch has $\lfloor l/2\rfloor+\lfloor l'/2\rfloor=(b-3)/2=\lfloor b/2\rfloor-1$.
\item $l,l'$ both even: the second branch has $-1+l/2+l'/2=\lfloor b/2\rfloor-1$.
\item Different parities: $b$ is even, and the first branch has $(b-2)/2=\lfloor b/2\rfloor-1$.
\end{itemize}

\emph{Case 2: $x$ in an $A$ gap of length $a$, so $l+l'=a-1$.} The two branches replace the gap by a $B$ gap and an $A$ gap, in the two possible orders, with the same $r$. Their contributions are
\[
\lfloor l/2\rfloor+\lceil l'/2\rceil\quad\text{and}\quad\lceil l/2\rceil+\lfloor l'/2\rfloor .
\]
\begin{itemize}[leftmargin=*]
\item $l$ or $l'$ even: one of them equals $\lfloor(a-1)/2\rfloor=\lceil a/2\rceil-1$.
\item $l,l'$ both odd: both equal $(a-1)/2=\lceil a/2\rceil-1$.
\end{itemize}

\emph{Case 3: $x$ in an $E$ gap of length $e$, with $l$ points between $x$ and its labelled neighbour.} The equal label gives a $B$ gap of length $l$ and an $E$ gap of length $e-1-l$, with the same $r$. The other label gives an $A$ gap of length $l$, an $E$ gap of length $e-1-l$, and spare budget $r-1$. Write $T=r+\sum_E e$; the first term of $\Phi$ is $\lfloor T/2\rfloor$.
\begin{itemize}[leftmargin=*]
\item $l$ even: in the second branch the first term and the new $A$ gap give $\lfloor (T-2-l)/2\rfloor+l/2=\lfloor T/2\rfloor-1$.
\item $l$ odd: in the first branch they give $\lfloor(T-1-l)/2\rfloor+(l-1)/2=\lfloor T/2\rfloor-1$.
\end{itemize}
All other terms are unchanged.
\end{proof}

For $F_k(x)$ there is a single gap of length $x-1$ ($B$ for even $k$ with $r=k$, $A$ for odd $k$ with $r=k-1$), so in both cases
\begin{equation}\label{eq:ub1}
L_k(x)\le\Bigl\lfloor\frac{x+k-1}{2}\Bigr\rfloor .
\end{equation}

\subsection{A block-pairing learner}

The learner below and the refinement move of Lemma~\ref{lem:moves}(e) are two sides of the same doubling step; Figure~\ref{fig:mechanism}(B) draws them together.

\begin{lemma}[Pairing]\label{lem:pair}
For $x\ge2$, $L_k(x)\le L_k(\lfloor x/2\rfloor)+k$.
\end{lemma}
\begin{proof}
\emph{Blocks.} Partition the points of $F_k(x)$ into \emph{blocks} of consecutive points: $\lfloor x/2\rfloor$ blocks for $R_k(x)$, and $\lfloor x/2\rfloor+1$ blocks for $G_k(x)$, whose two end blocks contain the fixed points. All blocks have size $2$ except at most one block of size $3$ or $1$, and no block contains two fixed points. Concretely, number the cycle of $R_k$ as $0,\dots,x-1$ with $0$ fixed, and the points of $G_k$ as $0,\dots,x$, and use the following blocks.
\begin{itemize}[leftmargin=*]
\item $R_k$, $x$ even: $\{0,1\},\{2,3\},\dots$.
\item $R_k$, $x$ odd: $\{x-1,0,1\},\{2,3\},\dots,\{x-3,x-2\}$.
\item $G_k$, $x$ even: $\{0,1\},\{2,3\},\dots,\{x-2,x-1\},\{x\}$.
\item $G_k$, $x$ odd: $\{0,1\},\dots,\{x-1,x\}$.
\end{itemize}
In each case the blocks, read in order, form an instance of $F_k(\lfloor x/2\rfloor)$ (a cycle of $\lfloor x/2\rfloor$ coarse points, or a path with $\lfloor x/2\rfloor$ coarse slots), whose fixed coarse points are the blocks that contain the fixed points.

\emph{The learner.} The learner keeps an optimal learner $A'$ for $F_k(\lfloor x/2\rfloor)$. The \emph{representative} of a block is its fixed point if it has one, and otherwise the first of its points to be presented. When a point $z$ of block $\beta$ arrives:
\begin{itemize}[leftmargin=*]
\item if $z$ becomes the representative of $\beta$, predict as $A'$ does on $\beta$, then pass $(\beta,f(z))$ to $A'$;
\item otherwise, predict the label of the representative of $\beta$.
\end{itemize}

\emph{Mistakes of the first kind.} The examples passed to $A'$ are consistent with the coarse function $g(\beta)=f(\mathrm{rep}(\beta))$. Since $g$ samples $f$ at points in order, it has no more switches than $f$, and it has the right fixed values. So $g$ lies in $F_k(\lfloor x/2\rfloor)$, and $A'$ errs at most $L_k(\lfloor x/2\rfloor)$ times.

\emph{Mistakes of the second kind.} A mistake at a non-representative $z$ means $f(z)\ne f(\mathrm{rep}(\beta))$, so a switch lies in a slot inside $\beta$ between $z$ and the representative. In a block of size $2$ this slot is unique and is charged once. The only block of size $3$ is $\{x-1,0,1\}$, with representative $0$, and its two possible mistakes are charged to the two different slots $(x-1,0)$ and $(0,1)$. Hence there are at most $k$ mistakes of the second kind.

The same charging can be stated for shattered trees: if $w$ slots are already known to hold a switch, then a tree of depth $D$ shattered by the current version space yields a tree of depth $D-(k-w)$ shattered by the coarse version space. The formal proof of Section~\ref{sec:verification} uses this form.
\end{proof}

\subsection{Adversary moves}

Call a shattered tree for a class \emph{full} if at every leaf the version space contains a function with exactly $k$ switches, and let $\Lambda_k(x)$ be the largest depth of a full tree for $F_k(x)$ ($-\infty$ if none exists). A function with exactly $k$ switches exists if and only if $x\ge k$, so $\Lambda_k(k)\ge0$. Every threshold function has exactly one switch, so $\Lambda_1=L_1=\lambda$ (Lemma~\ref{lem:thr}), and $\Lambda_0=L_0=0$.

\begin{lemma}\label{lem:moves}
Let $k\ge1$, and let $M$ stand for either $L$ or $\Lambda$, the same letter throughout each of items (b) to (d).
\begin{enumerate}[label=(\alph*)]
\item (Shattering) $L_k(x)=x-1$ for $1\le x\le k+1$.
\item (Monotonicity) $M_k(x+1)\ge M_k(x)$.
\item (Adjacent query) For $x\ge2$,
\[
M_k(x)\ge 1+\min\bigl(M_k(x-1),\,M_{k-1}(x-1)\bigr).
\]
\item (Query at distance two) For $x\ge3$,
\[
M_k(x)\ge 1+\min\bigl(M_k(x-2),\,1+M_{k-1}(x-2)\bigr).
\]
\item (Refinement) If $\Lambda_k(x)\ge0$, then
\[
L_k(2x)\ge\Lambda_k(x)+k\quad\text{and}\quad\Lambda_k(2x)\ge\Lambda_k(x)+k.
\]
\end{enumerate}
\end{lemma}
\begin{proof}
Let $z$ be the fixed point labelled $0$ and $y_1,y_2,\dots$ the free points following it.

\emph{(a) Shattering.} With $x-1\le k$ free points, every labelling has at most $x$ switches, one per slot. The parity of the switch count is fixed (even for $R_k$, odd for $G_k$), and $k$ has the same parity. So every labelling is allowed and the free points are shattered.

\emph{(b) Monotonicity.} Add a point next to $z$ and restrict to functions giving it label $0$. This subclass of $F_k(x+1)$ is isomorphic to $F_k(x)$, with the same switch counts.

\emph{(c) Adjacent query.} Query $y_1$.
\begin{itemize}[leftmargin=*]
\item Label $0$: merging $y_1$ with $z$ gives $F_k(x-1)$.
\item Label $1$: the slot $(z,y_1)$ carries a switch, and the remaining free points form a gap between $y_1$ (label $1$) and the far end, with budget $k-1$. For $R_k$ the far end is $z$ itself, reached by going around the cycle. Up to reversing the order of the points or complementing all labels, this is $G_{k-1}(x-1)$ if $k$ is even and $R_{k-1}(x-1)$ if $k$ is odd (for $R$, merge the two equally labelled ends into one fixed point). In both cases it is a copy of $F_{k-1}(x-1)$.
\end{itemize}
A function that uses all $k-1$ remaining switches uses all $k$ switches overall, so fullness is preserved.

\emph{(d) Query at distance two.} Query $y_2$.
\begin{itemize}[leftmargin=*]
\item Label $1$: the point $y_1$ lies between labels $0$ and $1$, so each of its two labels uses exactly one switch in the two slots around it, and the remaining points form $F_{k-1}(x-2)$ as in (c). The version space is the product of a two-element chain on $y_1$ and $F_{k-1}(x-2)$, of dimension $1+M_{k-1}(x-2)$ by Lemma~\ref{lem:prod}. Both labels of $y_1$ keep full leaves full.
\item Label $0$: the subclass with $y_1=0$ is isomorphic to $F_k(x-2)$.
\end{itemize}

\emph{(e) Refinement.} Insert one new point into every slot of $F_k(x)$. The result is $F_k(2x)$, in which the old points form a copy of $F_k(x)$.
\begin{enumerate}[label=\arabic*.,leftmargin=*]
\item Play a full tree of depth $\Lambda_k(x)$ on the old points. Every consistent labelling of the old points extends to the new ones (copy a neighbour), so the tree is shattered in $F_k(2x)$.
\item At a leaf, choose a function $g$ in the leaf's version space with exactly $k$ switches. Each switch of $g$ lies in a slot that now contains a new point, and that point may take either label without changing the number of switches. On the remaining new points, copy a neighbour.
\item This gives a product of $k$ two-element chains, of dimension $k$, whose functions all have exactly $k$ switches. Grafting these trees below the leaves gives a full tree of depth $\Lambda_k(x)+k$.\qedhere
\end{enumerate}
\end{proof}

\subsection{Properties of $Q_k$}

\begin{lemma}\label{lem:Q}
Let $k\ge1$ and $d_k=\lceil 3k/2\rceil$.
\begin{enumerate}[label=(\alph*)]
\item $Q_k$ is strictly increasing, and $Q_k(d)-Q_k(d-1)\ge2$ for $d\ge k+1$.
\item $Q_k(d)\ge 2d-k+1$ for all $d$; hence ${v_k(x)\le\lfloor(x+k-1)/2\rfloor}$.
\item $Q_k(e)\ge e+k$ for all $e\ge d_k+k$.
\item Let $k\ge2$ and $k+1\le d\le d_k+k+\lfloor k/2\rfloor-2$. Then
\[
Q_{k-1}(d-2)\le Q_k(d)-2.
\]
\end{enumerate}
\end{lemma}
\begin{proof}
(a) Within each line of the definition, $Q_k$ increases by $1$, by $2$ and (by induction) by at least $4$ per step. At $d=d_k+k$ the jump is
\[
2Q_k(d_k)-Q_k(d_k+k-1)=2d_k-3k+3\ge3.
\]

(b) Equality holds on the middle line, and the first line gives $d+1\ge2d-k+1$. On the third line $d\ge d_k+k\ge 5k/2$, so
\[
Q_k(d)=2Q_k(d-k)\ge2(2d-3k+1)\ge 2d-k+1.
\]

(c) For $e=d_k+k+i$ with $0\le i<k$ we have $Q_k(e)=2(2d_k+2i-k+1)$, and
\[
Q_k(e)\ge e+k\iff 3d_k+3i+2\ge4k,
\]
which holds. For larger $e$, by induction, $Q_k(e)=2Q_k(e-k)\ge2e\ge e+k$.

(d) Put $k'=k-1$ and $d_{k'}=\lceil 3k'/2\rceil$, so $d_k-d_{k'}$ equals $1$ for even $k$ and $2$ for odd $k$.

\emph{Middle line, $k+1\le d\le d_k+k-1$.} Here $Q_k(d)-2=2d-k-1$, and there are three possibilities for $d-2$.
\begin{itemize}[leftmargin=*]
\item $d-2\le k'$: then $Q_{k'}(d-2)=d-1\le2d-k-1$.
\item $d-2$ on the middle line for $k'$: then $Q_{k'}(d-2)=2d-k-2$.
\item Otherwise $k$ is odd and $d=d_k+k-1$, so $d-2=d_{k'}+k'$ and
\[
Q_{k'}(d-2)=2Q_{k'}(d_{k'})=4k-2=Q_k(d)-2.
\]
\end{itemize}

\emph{Third line, $d=d_k+k+i$ with $0\le i\le\lfloor k/2\rfloor-2$.} Here $Q_k(d)=2(2d_k+2i-k+1)$, and $d-2=d_{k'}+k'+i'$ with $i'=i$ for even $k$ and $i'=i+1$ for odd $k$. Since $i'\le k'-1$,
\[
Q_{k'}(d-2)=2(2d_{k'}+2i'-k'+1).
\]
Substituting $d_k=3k/2$ and $d_{k'}=3k/2-1$ (even $k$), or $d_k=(3k+1)/2$ and $d_{k'}=(3k-3)/2$ (odd $k$), gives $Q_{k'}(d-2)=Q_k(d)-2$ in both cases.
\end{proof}

\subsection{Proof of Theorem~\ref{thm:general}}

\emph{Upper bound.} By Lemma~\ref{lem:moves}(b), $L_k$ is nondecreasing, so it suffices to show $L_k(x)\le d$ at $x=Q_k(d+1)-1$. We use induction on $x$, and there are three regimes.
\begin{itemize}[leftmargin=*]
\item $d+1\le k$. Then $x=d+1$, and $L_k(x)\le x-1=d$ because there are $x-1$ free points.
\item $k\le d\le d_k+k-2$. Then $x=2d-k+2$, and \eqref{eq:ub1} gives $L_k(x)\le\lfloor(2d+1)/2\rfloor=d$.
\item $d+1\ge d_k+k$. Then $x=2Q_k(d+1-k)-1$, so $\lfloor x/2\rfloor=Q_k(d+1-k)-1$, and by induction $L_k(\lfloor x/2\rfloor)\le d-k$. Lemma~\ref{lem:pair} gives $L_k(x)\le d$.
\end{itemize}
The same argument gives $\Ldim\le v_k(x)$ for sequences with any fixed end labels and any $k\ge1$, without the parity condition stated at the start of this section.

\emph{Lower bound.} We prove, by induction on $k$ and then on $x$, that
\[
\begin{aligned}
&\text{(i)}\ \ L_k(x)\ge v_k(x),\\
&\text{(ii)}\ \Lambda_k(x)\ge\tau_k(x):=\min\bigl(x-k,\,v_k(x)\bigr)\ \ (x\ge k).
\end{aligned}
\]
For $k=1$ both hold with equality. Let $k\ge2$. By monotonicity it suffices to check (i) at $x=Q_k(d)$ for each $d\ge1$, and (ii) at the values $x>k$ where $\tau_k$ increases; for $x=k$, (ii) reads $\Lambda_k(k)\ge0$. Table~\ref{tab:moves} lists the move of Lemma~\ref{lem:moves} that settles each case.

\begin{table}[!htbp]
\centering
\caption{Adversary moves of Lemma~\ref{lem:moves} used in the lower-bound induction for Theorem~\ref{thm:general}; $d_k=\lceil 3k/2\rceil$.}
\label{tab:moves}
\begin{tabular}{|l|l|l|}
\hline
\textbf{Claim} & \textbf{Range} & \textbf{Move} \\ \hline
(i) at $x=Q_k(d)$ & $d\le k$ & (a) shattering \\ \hline
(i) at $x=Q_k(d)$ & $k+1\le d\le d_k+k+\lfloor k/2\rfloor-2$ & (d) query at distance two \\ \hline
(i) at $x=Q_k(d)$ & $d\ge d_k+k+\lfloor k/2\rfloor-1$ & (e) refinement \\ \hline
(ii), $\tau_k(x)=x-k$ & $x\le 3k-1$ & (c) adjacent query \\ \hline
(ii), $\tau_k(x)=v_k(x)$ & $x=Q_k(d)$, $d\ge k+1$ & (d) or (e), as for (i) \\ \hline
\end{tabular}
\end{table}

\emph{Claim (i) for $d\le k$.} This is Lemma~\ref{lem:moves}(a).

\emph{Claim (i) for $k+1\le d\le d_k+k+\lfloor k/2\rfloor-2$.} We use move (d). By Lemma~\ref{lem:Q}(a), $x-2\ge Q_k(d-1)$, and by Lemma~\ref{lem:Q}(d), $x-2\ge Q_{k-1}(d-2)$. Induction on $x$ and on $k$ gives
\[
L_k(x-2)\ge d-1,\qquad L_{k-1}(x-2)\ge d-2,
\]
and move (d) gives $L_k(x)\ge d$.

\emph{Claim (i) for larger $d$.} Then $d\ge d_k+k$ and $x=2Q_k(d-k)$ is even. Move (e) applies because $x/2=Q_k(d-k)\ge k+1$, so $\Lambda_k(x/2)\ge\tau_k(x/2)\ge0$ by (ii). It gives
\[
L_k(x)\ge\Lambda_k(x/2)+k\ge\tau_k\bigl(Q_k(d-k)\bigr)+k,
\]
and $\tau_k(Q_k(d-k))=\min(Q_k(d-k)-k,\,d-k)=d-k$ provided $Q_k(d-k)\ge d$. This condition holds in both remaining subcases.
\begin{itemize}[leftmargin=*]
\item $d-k=d_k+i$ with $\lfloor k/2\rfloor-1\le i<k$: the condition reads $2(d_k+i)-k+1\ge d_k+k+i$, that is, $i\ge 2k-1-d_k=\lfloor k/2\rfloor-1$.
\item $d-k\ge d_k+k$: the condition is Lemma~\ref{lem:Q}(c).
\end{itemize}

\emph{Claim (ii) when $\tau_k(x)=x-k$.} Then $x-k\le v_k(x)\le\lfloor(x+k-1)/2\rfloor$, so $x\le3k-1$. We use move (c) for $\Lambda$. Its two terms satisfy
\[
\begin{aligned}
\Lambda_k(x-1)&\ge\tau_k(x-1)\ge x-k-1,\\
\Lambda_{k-1}(x-1)&\ge\min\bigl(x-k,\,v_{k-1}(x-1)\bigr)\ge x-k-1.
\end{aligned}
\]
The first line holds because $v_k(x-1)\ge v_k(x)-1\ge x-k-1$, where the first inequality uses Lemma~\ref{lem:Q}(a): at most one threshold lies in $(x-1,x]$. For the second, $x-1\le 3k-2$ lies in the first two regimes of $Q_{k-1}$, which cover all sizes up to $Q_{k-1}(d_{k-1}+k-2)=2d_{k-1}+k-2\ge3k-2$ because $d_{k-1}\ge k$ for $k\ge2$; so
\[
v_{k-1}(x-1)=\min\bigl(x-2,\lfloor(x+k-3)/2\rfloor\bigr)\ge x-k-1,
\]
where the last inequality is $x\le3k-1$. Hence $\Lambda_k(x)\ge x-k$.

\emph{Claim (ii) when $\tau_k(x)=v_k(x)=d<x-k$.} Then $x=Q_k(d)$ with $d\ge k+1$. If $d\le d_k+k+\lfloor k/2\rfloor-2$, we use move (d) for $\Lambda$, with
\[
\begin{aligned}
\tau_k(x-2)&=\min\bigl(x-2-k,\,v_k(x-2)\bigr)\ge d-1,\\
\tau_{k-1}(x-2)&=\min\bigl(x-k-1,\,v_{k-1}(x-2)\bigr)\ge d-2.
\end{aligned}
\]
The first line holds because $x-k\ge d+1$, and the second by Lemma~\ref{lem:Q}(d). Otherwise we use move (e) exactly as in claim (i).

This completes the induction, and Theorem~\ref{thm:general} follows.\nobreak\hfill\qedsymbol

\begin{remark}
For $k=2$, Theorem~\ref{thm:general} and Lemma~\ref{lem:cyc} give $\Ldim(\Int_m)=v_2(m+1)$ with $Q_2=P$, a second proof of Theorem~\ref{thm:main}. The adversary of Section~\ref{sec:general} uses only two kinds of questions (next to a known $0$ and two steps from it) together with refinement, whereas the proof of Sections~\ref{sec:cut} and~\ref{sec:tree} describes every winning cut sequence.
\end{remark}

\begin{proof}[Proof of Corollary~\ref{cor:unions}]
The two identities follow from Theorem~\ref{thm:general}, Lemma~\ref{lem:cyc} and Theorem~\ref{thm:switch}.

For the estimate, let $x=m+1\ge4j+1=Q_{2j}(3j)$ and $d=v_{2j}(x)\ge 3j$. Every threshold with index $e=3j+2jt+i$ ($t\ge0$, $0\le i<2j$) satisfies
\[
\begin{gathered}
2j\lg\frac{Q_{2j}(e)}{j}-e=j+2j\,h_j\Bigl(\frac{i}{2j}\Bigr),\\
h_j(u)=\lg\Bigl(1+u+\frac1{4j}\Bigr)-u.
\end{gathered}
\]
On $[0,1)$ we have
\[
0\le\lg(1+u)-u\le h^\ast,
\]
where $h^\ast:=\max_u(\lg(1+u)-u)=-\lg\ln2-1/\ln2+1<0.0861$, and
\[
\lg\Bigl(1+u+\frac1{4j}\Bigr)\le\lg(1+u)+\frac{1}{4j\ln 2}.
\]
Hence $Q_{2j}(d)\le x$ gives $d\le 2j\lg(x/j)-j$, and $x<Q_{2j}(d+1)$ gives
\[
d+1>2j\lg\frac{x}{j}-j-2jh^\ast-\frac1{2\ln2},
\]
which is the stated lower bound.
\end{proof}

\begin{remark}
The upper bound in Corollary~\ref{cor:unions} is not attained: since $h_j(u)=\lg(1+u+\tfrac1{4j})-u>0$ on $[0,1)$, the slack is positive. Only the values $u=i/(2j)$ with $0\le i<2j$ occur, and on these the slack is at least $2\lg7-5\approx0.61$, attained at $j=1$, $i=1$. For $j=1$ the sharp form is Corollary~\ref{cor:asym}, with constant $2\lg7-4\approx1.615$ instead of $1$.
\end{remark}

\begin{proof}[Proof of Proposition~\ref{prop:odd}]
Let $V_p=\{f\in C_{n,k}:f(p)=0\}$. By complementation,
\[
\Ldim(C_{n,k})=1+\max_p\Ldim(V_p).
\] 

\emph{Lower bound.} For $p=1$, appending a virtual point labelled $1$ after position $n$ maps $V_1$ bijectively onto $G_k(n)$ without moving domain points. Indeed, a sequence that starts at $0$ and has $t$ switches ends in $1$ exactly when $t$ is odd, so the appended $1$ adds a switch exactly when $t$ is even; since $k$ is odd, an even $t\le k$ satisfies $t\le k-1$, and the total stays at most $k$. Theorem~\ref{thm:general} then gives $\Ldim(V_1)=v_k(n)$.

\emph{Upper bounds.} The class $V_p$ has two $E$ gaps of total length $n-1$ and spare budget $k$, so Lemma~\ref{lem:pot} gives $\lfloor(n+k-1)/2\rfloor$. Closing the path into a cycle adds at most one switch, and cyclic switch counts are even, so $V_p\subseteq R_{k+1}(n)$, and Theorem~\ref{thm:general} gives $\Ldim(V_p)\le v_{k+1}(n)$.

\emph{The case $k=1$.} Here $V_p$ consists of the constant $0$, the $n-p$ functions whose $1$-set is $[q,n]$ with $p<q\le n$, and the $p-1$ functions whose $1$-set is $[1,q]$ with $1\le q<p$. List their $1$-sets in the order
\[
[p+1,n],\ [p+2,n],\ \dots,\ [n,n],\ \emptyset,\ [1,1],\ [1,2],\ \dots,\ [1,p-1],
\]
that is, by the switch position read around the cycle starting at $p$. A query at $q>p$ answered $1$ keeps exactly the first $q-p$ functions of this list, and a query at $q<p$ answered $1$ keeps exactly the last $p-q$; so every query splits the list into a prefix and a suffix, every cut is available, and $\Ldim(V_p)=\lgf{n}$ as in Lemma~\ref{lem:thr}. Alternatively, $|C_{n,1}|=2n$, and a class that shatters a tree of depth $D$ has at least $2^D$ members, so $\Ldim(C_{n,1})\le\lgf{2n}=1+\lgf{n}$; the formal proof of Section~\ref{sec:verification} uses this form.
\end{proof}

\section{Implications for deterministic learners}\label{sec:remarks}

\begin{remark}[The natural learner is not optimal]\label{rem:natural}
For $m\ge1$, any learner that predicts $0$ on every point until it sees a positive example can be forced into
\[
N(m)=1+\max\{d:\ c(d)\le m+1\}
\]
mistakes. The adversary presents the point $x$ maximising $\lgf{x}+\lgf{m-x+1}$, answers ``positive'', and then plays the two threshold games of Lemma~\ref{lem:branch}(a). Binary search on both endpoints after the first positive matches this, so the natural learner makes exactly $N(m)$ mistakes in the worst case, and $\Ldim(\Int_m)\le N(m)$. In dimension one, this is the shape of the box learner of Chen and Maass \cite{chen1994line}.

From the explicit values, $P(d)\le c(d)\le P(d+1)$ for all $d\ge0$. Since
\[
m+1<P\bigl(\Ldim(\Int_m)+1\bigr)\le c\bigl(\Ldim(\Int_m)+1\bigr),
\]
we get $N(m)\le\Ldim(\Int_m)+1$, so
\[
\Ldim(\Int_m)\le N(m)\le\Ldim(\Int_m)+1.
\]
The gap is $1$ for $m=3,5,7,8,11,12,15,16,\dots$; among $1\le m<10^5$ it is $1$ for exactly $50{,}847$ values (about $51\%$; a finite count, not a limiting density). Hence the optimal learner must, in some states, predict the label $1$ before it has seen any positive example.
\end{remark}

\begin{remark}[Comparison with the counting bound]\label{rem:count}
The halving bound \cite{angluin1988queries,littlestone1988learning} $\lgf{|\Int_m|}=\lgf{1+m(m+1)/2}$ exceeds $\Ldim(\Int_m)$ by one for exactly $62{,}063$ of the values $1\le m<10^5$ (first at $m=5$) and never by more, so the exact value is not captured by counting.
\end{remark}

\begin{figure}[!htbp]
\centering
\includegraphics[width=\textwidth]{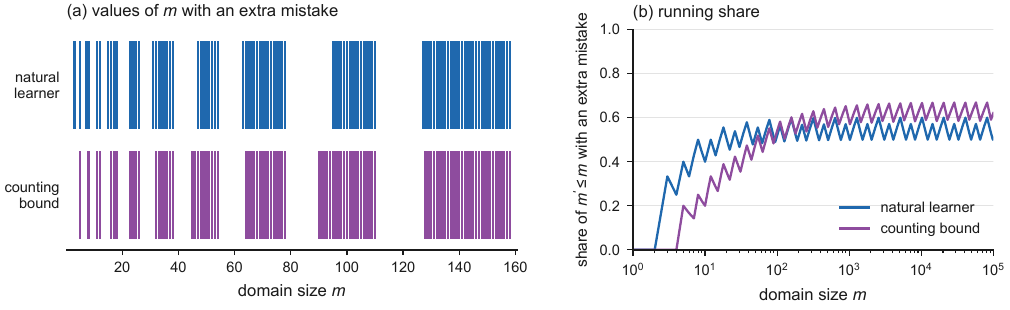}
\caption{The natural learner and the counting bound against the exact dimension. (a) The values $1\le m\le160$ at which $N(m)=\Ldim(\Int_m)+1$ (top) and $\lgf{|\Int_m|}=\Ldim(\Int_m)+1$ (bottom); at every other $m$ both quantities equal $\Ldim(\Int_m)$. (b) The share of $m'\le m$ with an extra mistake, for $m<10^5$; it ends at $50{,}847/99{,}999$ and $62{,}063/99{,}999$ (Remarks~\ref{rem:natural} and~\ref{rem:count}). All values are exact.}
\label{fig:gaps}
\end{figure}

Figure~\ref{fig:gaps}(a) marks the values $m\le160$ at which the natural learner or the counting bound is one above $\Ldim(\Int_m)$; both gaps occur in runs of consecutive $m$ that grow longer as $m$ grows. Panel (b) follows the share of such $m$ up to $10^5$.

\subsection{An explicit optimal learner}\label{sec:learner}

The upper-bound proof of Theorem~\ref{thm:general} gives a learner that can be run as it stands. On $R_k(x)$ or $G_k(x)$ with $x<Q_k(d_k+k-1)$ it predicts the label whose branch has the larger potential $\Phi$ of Lemma~\ref{lem:pot}, with ties going to $0$; when one branch is empty it predicts the only consistent label. By the proof of Lemma~\ref{lem:pot}, each mistake then lowers $\Phi$ by at least one. The initial potential is $\lfloor(x+k-1)/2\rfloor$ and there are $x-1$ free points, so in this base case the learner makes at most $\min(x-1,\lfloor(x+k-1)/2\rfloor)=v_k(x)$ mistakes. For larger $x$ it groups the points into the blocks of Lemma~\ref{lem:pair} and runs the same learner on the coarse instance of size $\lfloor x/2\rfloor$. The induction in the proof of Theorem~\ref{thm:general} then shows that it makes at most $v_k(x)$ mistakes: if $x\ge Q_k(d_k+k-1)$ and $d=v_k(x)$, then $d+1\ge d_k+k$, so $\lfloor x/2\rfloor<Q_k(d+1-k)$ and Lemma~\ref{lem:pair} adds $k$ mistakes to at most $d-k$. Intervals on $[m]$ are learned as $R_2(m+1)$, with point $0$ of the cycle fixed to $0$, and unions of $j$ intervals as $R_{2j}(m+1)$. For fixed $k$ the base case has constant size and the recursion has depth $O(\log x)$, so a round costs $O(\log x)$ dictionary operations, and the labels stored at all levels take $O(x)$ memory.

Against an exhaustive adaptive adversary the learner is forced into exactly $\Ldim(\Int_m)$ mistakes for $m\le10$, and into exactly $v_k(x)$ mistakes on $R_k(x)$ and $G_k(x)$ for $k\le6$ and $x\le10$ and on $G_3(x)$ for $x\le13$. These ranges reach the pairing step for $k=1,2,3$ (one level of pairing for $k=2,3$, several for $k=1$). The same search forces exactly $N(m)$ mistakes on the natural learner (Section~\ref{sec:verification}).

\begin{figure}[!htbp]
\centering
\includegraphics[width=0.6\textwidth]{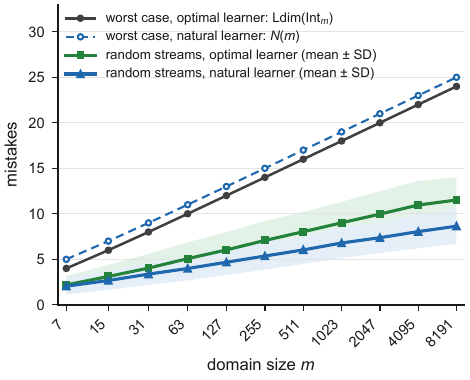}
\caption{Worst case and random streams for intervals on $m=2^t-1$ points, $3\le t\le13$. Upper lines: the exact worst case of the optimal learner, $\Ldim(\Int_m)$, and of the natural learner, $N(m)$ (Remark~\ref{rem:natural}); at these sizes $N(m)=\Ldim(\Int_m)+1$. Lower lines: mean number of mistakes, with a band of one population standard deviation, when the target is uniform over $\Int_m$ and the points arrive in uniformly random order (5000 trials per size for $m\le127$, down to 200 at $m=8191$; the values at $m=15,63,255,1023,4095$ are those of Table~\ref{tab:bench}).}
\label{fig:learners}
\end{figure}

In Table~\ref{tab:bench} the target is drawn uniformly from $\Int_m$, including the empty interval, and the points of $[m]$ arrive in uniformly random order; Figure~\ref{fig:learners} plots the same comparison for $m=2^t-1$, $3\le t\le13$, next to the two worst cases. The natural learner predicts $0$ until the first positive example. After it, the learner predicts $1$ between the smallest and the largest positive seen and $0$ where no remaining interval reaches; on either side of the positives it predicts $1$ exactly when more than half of the remaining candidates for that endpoint put the point inside the interval. On these streams the natural learner makes fewer mistakes on average than the optimal learner at every size tested, $8.04$ against $10.96$ at $m=4095$. Up to $m=255$ the optimal learner's largest count over the trials equals $\Ldim(\Int_m)$; the natural learner's reaches $\Ldim(\Int_m)$ only at $m=15$. In Python~3.14 on an Apple M4 Pro, the mean time per round, amortised over a whole stream, stays below one microsecond for both learners at every size.

\begin{table}[!htbp]
\centering
\caption{Mistakes of the optimal learner of Section~\ref{sec:learner} and the natural learner of Remark~\ref{rem:natural} on random streams: target uniform over $\Int_m$, points in uniformly random order. Mean (population standard deviation) and maximum over the trials; $\Ldim(\Int_m)$ is the worst case over all targets and orders.}
\label{tab:bench}
\footnotesize
\begin{tabular}{|r|r|r|c|r|c|r|}
\hline
\multirow{2}{*}{$m$} & \multirow{2}{*}{Trials} & \multirow{2}{*}{$\Ldim(\Int_m)$} & \multicolumn{2}{c|}{Optimal learner} & \multicolumn{2}{c|}{Natural learner} \\ \cline{4-7}
 & & & Mean (SD) & Max & Mean (SD) & Max \\ \hline
$15$ & $5000$ & $6$ & $3.12\ (1.28)$ & $6$ & $2.67\ (1.04)$ & $6$ \\ \hline
$63$ & $5000$ & $10$ & $5.07\ (1.79)$ & $10$ & $3.99\ (1.31)$ & $8$ \\ \hline
$255$ & $2000$ & $14$ & $7.07\ (2.12)$ & $14$ & $5.36\ (1.50)$ & $11$ \\ \hline
$1023$ & $1000$ & $18$ & $9.01\ (2.29)$ & $15$ & $6.79\ (1.67)$ & $12$ \\ \hline
$4095$ & $300$ & $22$ & $10.96\ (2.64)$ & $17$ & $8.04\ (1.83)$ & $12$ \\ \hline
\end{tabular}
\end{table}

\section{Randomised learners}\label{sec:random}

\begin{figure}[!htbp]
\centering
\includegraphics[width=0.6\textwidth]{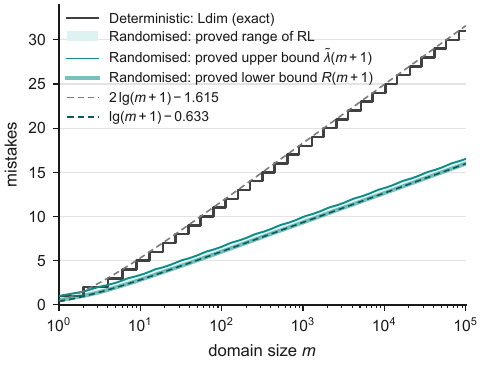}
\caption{Intervals on $m$ points: the exact deterministic dimension $\Ldim(\Int_m)$ (Theorem~\ref{thm:main}) and the randomised dimension, which lies in the shaded band between the proved lower bound $R(m+1)$ (Theorem~\ref{thm:rllower}) and the proved upper bound $\tilde\lambda(m+1)$ (Proposition~\ref{prop:rlupper}). The dashed lines are $2\lg(m+1)-(2\lg7-4)$, which the steps touch at $m+1=7\cdot2^{t-2}$ (Corollary~\ref{cor:asym}), and $\lg(m+1)-c^\ast$. All values are computed exactly from the closed forms.}
\label{fig:random}
\end{figure}

A randomised learner may predict a probability $p\in[0,1]$ of the label $1$ and pays $|p-y|$ in expectation. Filmus, Hanneke, Mehalel and Moran \cite{filmus2025optimal} showed that the optimal expected number of mistakes of a randomised learner in the realisable setting equals the \emph{randomised Littlestone dimension} $\mathrm{RL}(H)$, defined as half the largest expected length of a random root-to-leaf branch (each step left or right with probability $\tfrac12$) of a tree shattered by $H$. We determine $\mathrm{RL}$ exactly for thresholds, reduce $\mathrm{RL}(\Int_m)$ to an exact discounted cutting game, prove a lower bound that an explicit greedy adversary attains and an upper bound less than $0.64$ above it (Figure~\ref{fig:random}), and conjecture that the lower bound is exact; the conjecture holds for every $m\le70$.

For real $\ell\ge1$ write $\tilde\lambda(\ell)=t+(\ell-2^t)/2^t$ with $t=\lfloor\lg\ell\rfloor$, the piecewise-linear interpolation of $\lg$ between consecutive powers of two. It is concave on $[1,\infty)$ and satisfies $\tilde\lambda(2y)=\tilde\lambda(y)+1$.

\begin{lemma}\label{lem:rlrec}
Let $V$, $H_1$ and $H_2$ be finite nonempty classes, with $H_1$ and $H_2$ on disjoint domains.
\begin{enumerate}[label=(\alph*)]
\item (Recursion; Observation~5.5 of \cite{filmus2025optimal})
\[
\mathrm{RL}(V)=\max\Bigl(0,\ \max_x \tfrac12\bigl(1+\mathrm{RL}(V^0_x)+\mathrm{RL}(V^1_x)\bigr)\Bigr),
\]
where the inner maximum runs over points $x$ with $V^0_x,V^1_x\ne\emptyset$.
\item (Products) $\mathrm{RL}(H_1\otimes H_2)=\mathrm{RL}(H_1)+\mathrm{RL}(H_2)$.
\item (Invariance) $\mathrm{RL}$ is invariant under relabelling the domain and under deleting points on which the whole class agrees.
\end{enumerate}
\end{lemma}
\begin{proof}
\emph{Recursion.} A shattered tree is either a single leaf, with branch length $0$, or it has a root $x$ splitting $V$ and subtrees $T_0,T_1$ shattered by $V^0_x$ and $V^1_x$. In the second case the expected branch length is
\[
1+\tfrac12\bigl(E_{T_0}+E_{T_1}\bigr),
\]
and maximising each subtree separately gives the recursion.

\emph{Products.} Grafting an optimal tree for $H_2$ below every leaf of an optimal tree for $H_1$ adds the two expected branch lengths, which gives $\ge$. For $\le$, the learner runs optimal randomised learners for $H_1$ and $H_2$ on their own coordinates, and the characterisation of \cite{filmus2025optimal} bounds its expected loss by $\mathrm{RL}(H_1)+\mathrm{RL}(H_2)$.

\emph{Invariances.} Relabelling the domain and deleting points on which the whole class agrees map shattered trees to shattered trees of the same shape.
\end{proof}

\begin{proposition}\label{prop:rlthr}
A chain of $c$ threshold functions has $\mathrm{RL}=\tfrac12\tilde\lambda(c)$.
\end{proposition}
\begin{proof}
Let $t=\lgf{c}$; the case $c=1$ is $\mathrm{RL}=0=\tfrac12\tilde\lambda(1)$. Every split of a chain of $c$ functions leaves chains of $a$ and $b$ functions with $a+b=c$. By concavity and $\tilde\lambda(2y)=\tilde\lambda(y)+1$,
\[
\tilde\lambda(a)+\tilde\lambda(b)\le2\tilde\lambda(c/2)=2\tilde\lambda(c)-2,
\]
so by induction every split gives at most
\[
\tfrac12\Bigl(1+\tfrac12\tilde\lambda(a)+\tfrac12\tilde\lambda(b)\Bigr)\le\tfrac12\tilde\lambda(c).
\]
The split $a=\lfloor c/2\rfloor$, $b=\lceil c/2\rceil$ attains this bound, because both numbers lie in the same dyadic interval $[2^{t-1},2^t]$, on which $\tilde\lambda$ is linear.
\end{proof}

\paragraph{Intervals as a discounted cutting game.} Using the segment states $J_S$ of Section~\ref{sec:cut} with pieces of measure $\ell=g+1$: the proof of Lemma~\ref{lem:branch}(a) shows that the positive branch of a query is the product of two chains of $u$ and $v$ functions with $u+v=\ell$, and part (b) gives the negative branch. With Lemma~\ref{lem:rlrec} and Proposition~\ref{prop:rlthr} this yields, exactly,
\begin{equation}\label{eq:rlcut}
\begin{aligned}
\mathrm{RL}(J_S)=\max\Bigl(0,\ \max_{\text{cuts } \ell=u+v}\ \tfrac12\Bigl(&1+\tfrac12\tilde\lambda(u)+\tfrac12\tilde\lambda(v)\\
&+\mathrm{RL}(J_{S'})\Bigr)\Bigr).
\end{aligned}
\end{equation}
Unrolled, the adversary chooses a sequence of cuts and collects
\[
\sum_i 2^{-i}\Bigl(1+\tfrac12\tilde\lambda(u_i)+\tfrac12\tilde\lambda(v_i)\Bigr).
\]
This is the deterministic cutting game of Section~\ref{sec:cut} with two changes: \emph{discounting} replaces the constraint ``value $\ge$ remaining depth'', and the cut value $\tfrac12(\tilde\lambda(u)+\tilde\lambda(v))$ replaces $\lgf u+\lgf v$.

We evaluated \eqref{eq:rlcut} exactly, in rational arithmetic, for $m\le70$ (Section~\ref{sec:verification}); for $m\le12$ it agrees with the generic recursion on explicit function tables. For $m=0,1,2,\dots$ the values start
\[
0,\ \frac12,\ 1,\ \frac{11}{8},\ \frac{27}{16},\ \frac{63}{32},\ \frac{139}{64},\ \frac{303}{128},\ \dots
\]

Optimal play is far from unique. For $m=16$, the first cut of the rod $17$ can be $8+9$, $5+12$ or $4+13$, and one optimal line continues by peeling short blocks off the long piece ($13=4+9$, $9=2+7$, $7=2+5$, $5=2+3$). The choice among cuts of equal immediate value matters: always cutting a largest piece as evenly as possible gives $\tfrac{61}{32}$ instead of $\tfrac{63}{32}$ for $m=5$ (rod $6$ cut as $3+3$ rather than $2+4$). 

A simple rule does reach the optimum in all computed cases: cut a largest piece $\ell$, with $t=\lgf{\ell}$, into
\[
u=\max(2^{t-1},\,\ell-2^t),\qquad v=\ell-u.
\]
Every cut of this rule has the largest possible immediate value $\tfrac12(\tilde\lambda(u)+\tilde\lambda(v))=\tilde\lambda(\ell)-1$, and each cut produces a piece whose measure is a power of two. Theorem~\ref{thm:rllower} below evaluates this strategy exactly for every $m$; its value equals $\mathrm{RL}(\Int_m)$ for every $m\le70$.

Let $R(1)=0$ and $R(\ell)=1-2^{1-\ell}+\tfrac12\bigl(R(\lfloor\ell/2\rfloor)+R(\lceil\ell/2\rceil)\bigr)$ for $\ell\ge2$.

\begin{lemma}\label{lem:Rclosed}
For $\ell\ge1$, with $L=\lgf{\ell}$, $f_j=\lfloor\ell/2^j\rfloor$ and $\pi_j=\ell/2^j-f_j$,
\[
R(\ell)=\sum_{j=0}^{L}\Bigl(1-2^{1-f_j}+\pi_j\,2^{-f_j}\Bigr).
\]
\end{lemma}
\begin{proof}
Write $g(x)=1-2^{1-x}$, so that $g(1)=0$ and
\[
R(\ell)=g(\ell)+\tfrac12\bigl(R(\lfloor\ell/2\rfloor)+R(\lceil\ell/2\rceil)\bigr).
\]
Consider the random process $X_0=\ell$, $X_{j+1}\in\{\lfloor X_j/2\rfloor,\lceil X_j/2\rceil\}$, each with probability $\tfrac12$.
\begin{enumerate}[label=\arabic*.,leftmargin=*]
\item By induction $X_j\in\{f_j,f_j+1\}$. So $X_j\ge f_j\ge2$ for $j<L$, and $X_L\in\{1,2\}$, where $R(X_L)=g(X_L)$. Unrolling the recurrence therefore gives
\[
R(\ell)=\sum_{j=0}^{L}\mathbb{E}[g(X_j)].
\]
\item The identity $\mathbb{E}[X_{j+1}\mid X_j]=X_j/2$ gives $\mathbb{E}[X_j]=\ell/2^j$, so $\Pr[X_j=f_j+1]=\pi_j$.
\item Hence
\[
\mathbb{E}[g(X_j)]=(1-\pi_j)\,g(f_j)+\pi_j\, g(f_j+1)=1-2^{1-f_j}+\pi_j2^{-f_j}.
\]
\end{enumerate}
\end{proof}

\begin{theorem}\label{thm:rllower}
For every $m\ge0$, $\mathrm{RL}(\Int_m)\ge R(m+1)$.
\end{theorem}
\begin{proof}
Let $\ell=m+1$. By \eqref{eq:rlcut}, every cut $\ell=u+v$ with value $w=\tfrac12(\tilde\lambda(u)+\tilde\lambda(v))$ gives
\[
\mathrm{RL}(J_S)\ge\tfrac12\bigl(1+w+\mathrm{RL}(J_{S'})\bigr),
\]
so any complete schedule of cuts gives a lower bound equal to its discounted value $\sum_i2^{-i}(1+w_i)$. Consider the greedy schedule: always cut a largest piece $p$, with $t=\lgf p$, into
\[
u=\max(2^{t-1},\,p-2^t),\qquad v=p-u.
\]

\emph{Step 1: each cut has value $1+w=\tilde\lambda(p)$.} If $p\le3\cdot2^{t-1}$, then $u=2^{t-1}$ and $v\in[2^{t-1},2^t]$; otherwise $v=2^t$ and $u\in(2^{t-1},2^t)$. In both cases a direct computation on the linear pieces of $\tilde\lambda$ gives $\tfrac12(\tilde\lambda(u)+\tilde\lambda(v))=\tilde\lambda(p)-1$. Each cut piece is a largest current piece and its children are smaller, so the pieces are cut in non-increasing order of size. Hence the value of the schedule is
\[
\sum_i2^{-i}\,\tilde\lambda(p_i),
\]
where $p_1\ge p_2\ge\dots\ge p_{\ell-1}$ are the pieces that are cut, in the order they are cut.

\emph{Step 2: level counts.} Let $C_q$ be the number of cut pieces of measure at least $2^q$. We claim
\[
C_q=f_{q-1}-1\qquad(1\le q\le L+1).
\]
For a piece $p\ge2^{q-1}$ let $N_q(p)$ be the number of cut pieces of measure at least $2^q$ among $p$ and its descendants; then $C_q=N_q(\ell)$, and we show $N_q(p)=\lfloor p/2^{q-1}\rfloor-1$ by induction on $p$. A piece in $[2^{q-1},2^q)$ contributes nothing, which matches $\lfloor p/2^{q-1}\rfloor-1=0$. If $p\ge2^q$, then $t\ge q$, the piece is cut, and both children have size at least $2^{t-1}\ge2^{q-1}$. One of $u,v$ is a power of two at least $2^{t-1}$, hence divisible by $2^{q-1}$, so $\lfloor u/2^{q-1}\rfloor+\lfloor v/2^{q-1}\rfloor=\lfloor p/2^{q-1}\rfloor$. Therefore
\[
\begin{aligned}
N_q(p)&=1+N_q(u)+N_q(v)\\
&=1+\bigl(\lfloor u/2^{q-1}\rfloor-1\bigr)+\bigl(\lfloor v/2^{q-1}\rfloor-1\bigr)\\
&=\lfloor p/2^{q-1}\rfloor-1 .
\end{aligned}
\]

\emph{Step 3: the spine.} Every cut produces a power of two and a remainder, and powers of two split evenly. By induction, the pieces that are not powers of two are exactly
\[
s_t=2^t+(\ell\bmod2^t)
\]
for the levels $1\le t\le L$ with $\ell\bmod2^t\ne0$, one per level. The rod itself is $s_L$ unless $\ell$ is a power of two, and if $s_t=2^t+r$ with $r=\ell\bmod2^t\in(0,2^t)$, the greedy cut gives
\begin{itemize}[leftmargin=*]
\item for $r\le2^{t-1}$, the pieces $2^{t-1}$ and $2^{t-1}+r$;
\item for $r>2^{t-1}$, the pieces $2^t$ and $r=2^{t-1}+(r-2^{t-1})$.
\end{itemize}
In both cases the piece that is not a power of two, if there is one, equals $2^{t-1}+(\ell\bmod2^{t-1})=s_{t-1}$. Since $s_t$ is the largest piece with size in $[2^t,2^{t+1})$, it is cut in position $C_{t+1}+1=f_t$, and $\tilde\lambda(s_t)-t=\pi_t$.

\emph{Step 4: the total.} Write $\tilde\lambda(p)=\lgf p+(\tilde\lambda(p)-\lgf p)$. Because the pieces are cut in non-increasing order,
\[
\sum_i2^{-i}\lgf{p_i}=\sum_{q=1}^{L}\bigl(1-2^{-C_q}\bigr)=\sum_{j=0}^{L-1}\bigl(1-2^{1-f_j}\bigr),
\]
and by Step 3 the fractional parts contribute $\sum_t 2^{-f_t}\pi_t$. The $j=L$ term of Lemma~\ref{lem:Rclosed} has $f_L=1$, and $\pi_0=0$, so the total is exactly $R(\ell)$.
\end{proof}

\begin{conjecture}\label{conj:rl}
For every position $J_S$, $\mathrm{RL}(J_S)$ equals the value of the greedy schedule started from $S$. In particular $\mathrm{RL}(\Int_m)=R(m+1)$ for all $m\ge0$.
\end{conjecture}

The identity $\mathrm{RL}(\Int_m)=R(m+1)$ was verified exactly, in rational arithmetic, for every $m\le70$, and the greedy schedule attains the exact optimum of \eqref{eq:rlcut} on all $5{,}603$ multisets of pieces of measure at least $2$ with total measure at most $30$. Since greedy play is one available strategy, the conjecture is equivalent to the statement that the greedy value is a supersolution of \eqref{eq:rlcut}: from every position, one arbitrary cut followed by greedy play is never better than greedy play.

The lower bound satisfies $R(2\ell)=R(\ell)+1-2^{1-2\ell}$, so it grows by exactly $1-2^{1-2\ell}$ when the domain measure doubles; this is the randomised counterpart of Theorem~\ref{thm:general}, where doubling adds $k=2$ mistakes. At powers of two, $R(2^t)=t-\sum_{s=1}^t2^{1-2^s}$, so
\[
\mathrm{RL}(\Int_{2^t-1})\ge R(2^t)>t-c^\ast,
\]
\[
c^\ast=\sum_{s\ge1}2^{1-2^s}=0.632843018\ldots,
\]
with equality in the first inequality under the conjecture; the offset $t-R(2^t)$ increases to $c^\ast$, twice a lacunary series, and numerically $\lg(m+1)-0.6428<R(m+1)<\lg(m+1)-0.6162$ for $3\le m<2\cdot10^5$. In particular a randomised learner for $\Int_m$ cannot do better than $\lg(m+1)-0.6428$ expected mistakes in this range.

Without the terms $2^{1-\ell}$ the recurrence would give exactly $\tilde\lambda(\ell)$, so the theorem and conjecture together read
\[
\mathrm{RL}(\Int_m)=\tilde\lambda(m+1)-\delta(m+1),
\]
with a correction $\delta$ that satisfies $\delta(1)=0$ and
\[
\delta(\ell)=2^{1-\ell}+\tfrac12\bigl(\delta(\lfloor\ell/2\rfloor)+\delta(\lceil\ell/2\rceil)\bigr).
\]
Since randomised learners include deterministic ones, $\mathrm{RL}(H)\le\Ldim(H)$ for every class $H$, which for intervals gives only $\mathrm{RL}(\Int_m)\le2\lg(m+1)-1.61$ when $m\ge3$ (Corollary~\ref{cor:asym}). The following bound needs neither the conjecture nor the deterministic results, and it leaves a gap of less than $c^\ast$.

\begin{proposition}\label{prop:rlupper}
For every position $J_S$, with pieces of measure $\ell=g+1$ as in \eqref{eq:rlcut}, $\mathrm{RL}(J_S)\le\max_{\ell}\tilde\lambda(\ell)$, the maximum taken over the pieces and equal to $0$ if there are none. In particular, for every $m\ge0$,
\[
\tilde\lambda(m+1)-c^\ast<R(m+1)\le\mathrm{RL}(\Int_m)\le\tilde\lambda(m+1)\le\lg(m+1).
\]
More precisely, $\tilde\lambda(\ell)-R(\ell)\le c_L=\sum_{s=1}^{L}2^{1-2^s}$ for $2^L\le\ell<2^{L+1}$, with equality at $\ell=2^L$, so the constant $c^\ast=\lim_L c_L$ in the first inequality cannot be improved.
\end{proposition}
\begin{proof}
\emph{Upper bound.} Write $f(S)$ for this maximum and induct on the sum of $\ell-1$ over the pieces, which decreases with every cut. If no piece can be cut, $\mathrm{RL}(J_S)=0\le f(S)$. Otherwise consider a cut $\ell=u+v$ of a piece of measure $\ell$, leading to $S'$. Since $\tilde\lambda$ is concave and $\tilde\lambda(2y)=\tilde\lambda(y)+1$, as in the proof of Proposition~\ref{prop:rlthr},
\[
\tfrac12\bigl(\tilde\lambda(u)+\tilde\lambda(v)\bigr)\le\tilde\lambda(\ell/2)=\tilde\lambda(\ell)-1\le f(S)-1 .
\]
Every piece of $S'$ is at most a piece of $S$, so $\mathrm{RL}(J_{S'})\le f(S')\le f(S)$ by induction, and in \eqref{eq:rlcut} the cut is worth at most
\[
\tfrac12\bigl(1+(f(S)-1)+f(S)\bigr)=f(S).
\] As $f(S)\ge0$, $\mathrm{RL}(J_S)\le f(S)$. For $\Int_m=J_{\{m\}}$, a single piece of measure $m+1$, this gives $\mathrm{RL}(\Int_m)\le\tilde\lambda(m+1)$, and $\tilde\lambda\le\lg$ because $\tilde\lambda$ interpolates the concave function $\lg$ by chords.

\emph{Gap.} The middle inequality is Theorem~\ref{thm:rllower}. For the first, let $\delta(\ell)=\tilde\lambda(\ell)-R(\ell)$. For $\ell\ge2$ the halves $\lfloor\ell/2\rfloor$ and $\lceil\ell/2\rceil$ lie in one interval $[2^{t},2^{t+1}]$ on which $\tilde\lambda$ is linear, so $\tilde\lambda(\ell)=1+\tfrac12\bigl(\tilde\lambda(\lfloor\ell/2\rfloor)+\tilde\lambda(\lceil\ell/2\rceil)\bigr)$; subtracting the recurrence for $R$ gives $\delta(1)=0$ and
\[
\delta(\ell)=2^{1-\ell}+\tfrac12\bigl(\delta(\lfloor\ell/2\rfloor)+\delta(\lceil\ell/2\rceil)\bigr).
\]
We show $\delta(\ell)\le c_L$ for $2^L\le\ell<2^{L+1}$ by induction on $\ell$; the case $\ell=1$ is $\delta(1)=0=c_0$. Let $\ell\ge2$. If both halves are below $2^L$, they lie in $[2^{L-1},2^L)$, so $\delta(\ell)\le2^{1-\ell}+c_{L-1}\le2^{1-2^L}+c_{L-1}=c_L$. Otherwise $\lceil\ell/2\rceil=2^L$ and $\lfloor\ell/2\rfloor=2^L-1$, so $\ell=2^{L+1}-1$, and $\delta(2^L)\le c_L$ by induction; then $\delta(\ell)\le2^{1-\ell}+\tfrac12(c_{L-1}+c_L)=c_L+2^{1-\ell}-2^{-2^L}\le c_L$, because $\ell-1\ge2^L$ for $L\ge1$. For $\ell=2^L$ both halves equal $2^{L-1}$, so $\delta(2^L)=2^{1-2^L}+\delta(2^{L-1})=c_L$ by induction. Since $c_L<c^\ast$, $R(m+1)>\tilde\lambda(m+1)-c^\ast$.
\end{proof}

A simple learner achieves the upper bound: its expected number of mistakes is at most $\tilde\lambda(m+1)$. Predicting $0$ until the first positive example costs at most one mistake, since every earlier example is negative. After a positive example inside a block of $g\le m$ unlabelled points, the possible left and right endpoints form two chains of sizes $u$ and $v$ with $u+v=g+1\le m+1$, and optimal randomised threshold learners for the two chains (Proposition~\ref{prop:rlthr}) add at most $\tfrac12\tilde\lambda(u)+\tfrac12\tilde\lambda(v)\le\tilde\lambda(m+1)-1$ expected mistakes; the labels of all other points are forced. Proposition~\ref{prop:rlupper} gives $\lg(m+1)-0.72<\mathrm{RL}(\Int_m)\le\lg(m+1)$ for every $m$, using $\tilde\lambda(\ell)\ge\lg\ell-0.0861$, so randomisation halves the leading term $2\lg(m+1)$ of the deterministic bound. At $m+1=2^t$ the two bounds are $t-c_t$ and $t$, so the remaining uncertainty there is below $c^\ast$, and Conjecture~\ref{conj:rl} would remove it.

Figure~\ref{fig:random} compares the two randomised bounds with the deterministic dimension $\Ldim(\Int_m)$ for $m\le10^5$; the band between them grows at half the deterministic rate.

\section{Computational verification}\label{sec:verification}

Every statement was also checked by exact computation; Table~\ref{tab:verification} lists the checks and their ranges. The first row uses only the recursion of Fact~\ref{fact:rec} on explicit function tables, with no structural lemma; this generic computation takes quasi-polynomial time in general \cite{manurangsi2017inapproximability}, so it covers only small ranges, and a second implementation written separately repeats it on larger ranges. Randomised values are computed in rational arithmetic. All checks agree with the statements. The code that reproduces them is public under the MIT license at \url{https://github.com/researchbackup27-sudo/exact-littlestone-intervals} \cite{daga2026code}, together with a script that reruns every computational row of Table~\ref{tab:verification}.

\begin{table}[!htbp]
\centering
\caption{Exact computations that check each statement, with the ranges covered.}
\label{tab:verification}
\footnotesize
\begin{tabularx}{\textwidth}{|p{3.6cm}|X|p{5.2cm}|}
\hline
\textbf{Statement} & \textbf{Check} & \textbf{Range} \\ \hline
Theorems~\ref{thm:main} and~\ref{thm:switch} & Recursion of Fact~\ref{fact:rec} on explicit classes; second implementation & $\Int_m$: $m\le12$ ($m\le16$); $C_{n,2}$: $n\le11$ ($n\le13$) \\ \hline
Theorems~\ref{thm:main}, \ref{thm:switch} and~\ref{thm:general}; identities of Corollary~\ref{cor:unions}; Proposition~\ref{prop:odd} & Formal proof in Lean~4 with Mathlib, checked by the Lean kernel & every $m$; every $k\ge1$, $x\ge1$; every $n\ge1$, $j\ge1$; every $n\ge1$, odd $k$ \\ \hline
Theorems~\ref{thm:switch} and~\ref{thm:general}, Proposition~\ref{prop:odd} & Compressed dynamic program over labelled samples of $C_{n,k}$ & $k=2$: $n\le28$; $k\le4$: $n\le11$ \\ \hline
Theorem~\ref{thm:switch} & Reduction to $U_j(n-1)$ & $k=2,4,6$; $n\le11$ \\ \hline
Lemma~\ref{lem:cut} & Segment recursion & $m\le60$ \\ \hline
Propositions~\ref{prop:upper} and~\ref{prop:lower} & Minimum root measure of admissible trees & $d\le19$ ($m+1\le1280$) \\ \hline
Lemma~\ref{lem:elem}(b) & Formula for $g$ against brute force & $a,b<40$; $r\le8$ \\ \hline
Theorem~\ref{thm:main}, Remarks~\ref{rem:natural} and~\ref{rem:count} & Closed form; the two counts & closed form: $3\le m<10^5$; counts: $1\le m<10^5$ \\ \hline
Theorem~\ref{thm:general}, Corollary~\ref{cor:unions} & Exact solver for the cyclic gap game, against the recursion of Fact~\ref{fact:rec} and against $v_{2j}$ & $U_2,U_3$: $m\le11$; $j=1$: $m\le40$; $j=2$: $m\le45$; $j=3$: $m\le28$; single gaps against $v_{2s+1}$: $s=0,1,2,3$ with $x\le40,30,22,18$ \\ \hline
Theorem~\ref{thm:general} & Upper-bound recursion $\min(x-1,\lfloor(x+k-1)/2\rfloor,L(\lfloor x/2\rfloor)+k)$ equals $v_k$ & $k\le16$; $x\le2\cdot10^5$ \\ \hline
Theorem~\ref{thm:general} & Lower-bound system of Lemma~\ref{lem:moves} equals $v_k$; case analysis of the induction replayed move by move & $k\le25$, $x\le3000$; case analysis $k\le40$ \\ \hline
Proposition~\ref{prop:odd} & $\Ldim(C_{n,k})=1+v_k(n)$ & $k=1$: $n\le40$; $k=3$: $n\le26$; $k=5$: $n\le18$; $k=7$: $n\le15$ \\ \hline
Lemma~\ref{lem:rlrec}, Proposition~\ref{prop:rlthr} & Randomised recursion on explicit classes & chains $c\le17$; $\Int_m$: $m\le12$; products of chains $a,b\le6$ \\ \hline
Equation~\eqref{eq:rlcut}, Conjecture~\ref{conj:rl} & Discounted cutting game solved exactly and compared with $R(m+1)$ & $m\le70$ \\ \hline
Lemma~\ref{lem:Rclosed}, Theorem~\ref{thm:rllower} & Closed form, level counts, spine and value of the greedy schedule; greedy rule against the exact solver and against $R(m+1)$ & $\ell<400$; solver: $m\le30$; $R(m+1)$: $m<600$ \\ \hline
Conjecture~\ref{conj:rl} & Greedy schedule against the exact optimum from every position & all $5{,}603$ multisets of pieces of measure at least $2$ with total measure at most $30$ \\ \hline
Bounds on $R(m+1)$ & $\lg(m+1)-0.6428<R(m+1)<\lg(m+1)-0.6162$ & $3\le m<2\cdot10^5$ \\ \hline
Proposition~\ref{prop:rlupper} & Upper bound $\max_{\ell}\tilde\lambda(\ell)$ against the exact solver; $\mathrm{RL}(\Int_m)\le\tilde\lambda(m+1)$; $\tilde\lambda(\ell)-R(\ell)\le c_L$, with equality at $\ell=2^L$ & all $2{,}435$ multisets of pieces of measure at least $2$ with total measure at most $26$; $m\le70$; gap: exact for $\ell\le2048$, floating point for $\ell\le2\cdot10^5$ \\ \hline
Lemmas~\ref{lem:pot} and~\ref{lem:pair}, Theorems~\ref{thm:main} and~\ref{thm:general}, Remark~\ref{rem:natural} & Learner of Section~\ref{sec:learner} and natural learner against an exhaustive adaptive adversary; mistake bound of the learner's recursion against $v_k$ & $\Int_m$: $m\le10$; $R_k$, $G_k$: $k\le6$, $x\le10$; $G_3$: $x\le13$; recursion: $k\le16$, $x\le2\cdot10^5$ \\ \hline
\end{tabularx}
\end{table}

\begin{figure}[!htbp]
\begin{lstlisting}
abbrev Hyp (m : ℕ) := Fin m → Bool

def restrict (H : Finset (Hyp m)) (x : Fin m) (b : Bool) : Finset (Hyp m) :=
  H.filter (fun h => h x = b)

def Shatters : Finset (Hyp m) → ℕ → Prop
  | H, 0 => H.Nonempty
  | H, d + 1 => ∃ x : Fin m, Shatters (restrict H x false) d ∧ Shatters (restrict H x true) d

def ldim (H : Finset (Hyp m)) : ℕ :=
  Nat.findGreatest (fun d => Shatters H d) m

def interval (a b : ℕ) : Hyp m := fun x => decide (a ≤ x.val ∧ x.val ≤ b)

def intervals (m : ℕ) : Finset (Hyp m) :=
  insert (fun _ => false)
    (((Finset.univ : Finset (Fin m × Fin m)).filter (fun p => p.1 ≤ p.2)).image
      (fun p => interval p.1.val p.2.val))

def P : ℕ → ℕ
  | 0 => 1
  | 1 => 2
  | 2 => 3
  | 3 => 5
  | d + 4 => if d % 2 = 0 then 7 * 2 ^ (d / 2) else 5 * 2 ^ ((d - 1) / 2 + 1)

theorem main (m d : ℕ) : Shatters (intervals m) d ↔ P d ≤ m + 1 := …

theorem ldim_intervals_isGreatest (m : ℕ) :
    IsGreatest {d | P d ≤ m + 1} (ldim (intervals m)) := …

theorem ldim_intervals_closed_form (m : ℕ) (hm : 3 ≤ m) :
    ldim (intervals m) =
      max (2 * Nat.log 2 (4 * (m + 1) / 7)) (2 * Nat.log 2 (2 * (m + 1) / 5) + 1) := …

#print axioms main  -- 'ExactLittlestone.main' depends on axioms: [propext, Classical.choice, Quot.sound]
\end{lstlisting}
\caption{Lean~4 definitions and statements of Theorem~\ref{thm:main}, as in the files \texttt{Defs.lean}, \texttt{Main.lean} and \texttt{Paper.lean} of the repository. The namespace \texttt{ExactLittlestone}, the implicit arguments \texttt{\{m : $\mathbb{N}$\}} and the decidability instance for \texttt{Shatters} (which \texttt{Nat.findGreatest} requires and which does not affect its value) are omitted, and proofs are replaced by \texttt{\dots}; the last line is the output of the axiom check in \texttt{Check.lean}.}
\label{fig:leancode}
\end{figure}

\begin{figure}[!htbp]
\begin{lstlisting}
def switches : List Bool → ℕ
  | a :: b :: l => (if a = b then 0 else 1) + switches (b :: l)
  | _ => 0

def pathClass (a b : Bool) (k n : ℕ) : Finset (Hyp n) :=
  Finset.univ.filter (fun h => switches (a :: List.ofFn h ++ [b]) ≤ k)

def cyc (k x : ℕ) : Finset (Hyp (x - 1)) := pathClass false false k (x - 1)

def gap (k x : ℕ) : Finset (Hyp (x - 1)) := pathClass false true k (x - 1)

def switchSeq (n k : ℕ) : Finset (Hyp n) :=
  Finset.univ.filter (fun h => switches (List.ofFn h) ≤ k)

def unions (j m : ℕ) : Finset (Hyp m) :=
  Finset.univ.filter (fun h =>
    ∃ S ∈ (intervals m).powerset, S.card ≤ j ∧ ∀ x, h x = true ↔ ∃ I ∈ S, I x = true)

def dk (k : ℕ) : ℕ := (3 * k + 1) / 2

def Q (k d : ℕ) : ℕ :=
  if d ≤ k then d + 1
  else if d < dk k + k then 2 * d + 1 - k
  else if k = 0 then 0
  else 2 * Q k (d - k)
termination_by d
decreasing_by omega

def v (k x : ℕ) : ℕ := Nat.findGreatest (fun d => Q k d ≤ x) x

theorem general (k x : ℕ) (hk : 1 ≤ k) (hx : 1 ≤ x) :
    (Even k → ldim (cyc k x) = v k x) ∧ (Odd k → ldim (gap k x) = v k x) := …

theorem shatters_pathClass_iff (hk : 1 ≤ k) (hpar : k % 2 = if a = b then 0 else 1) :
    Shatters (pathClass a b k n) d ↔ Q k d ≤ n + 1 := …

theorem ldim_switchSeq_even {n j : ℕ} (hn : 1 ≤ n) :
    ldim (switchSeq n (2 * j)) = 1 + ldim (unions j (n - 1)) := …

theorem ldim_unions {j : ℕ} (hj : 1 ≤ j) (m : ℕ) : ldim (unions j m) = v (2 * j) (m + 1) := …

theorem ldim_switchSeq {j n : ℕ} (hj : 1 ≤ j) (hn : 1 ≤ n) :
    ldim (switchSeq n (2 * j)) = 1 + v (2 * j) n := …

theorem prop6_lower (hn : 1 ≤ n) (hk : Odd k) : 1 + v k n ≤ ldim (switchSeq n k) := …

theorem prop6_upper (hn : 1 ≤ n) (hk : Odd k) :
    ldim (switchSeq n k) ≤ 1 + min ((n + k - 1) / 2) (v (k + 1) n) := …

theorem prop6_one (hn : 1 ≤ n) : ldim (switchSeq n 1) = 1 + Nat.log 2 n := …

#print axioms general  -- 'ExactLittlestone.general' depends on axioms: [propext, Classical.choice, Quot.sound]
\end{lstlisting}
\caption{Lean~4 definitions and statements of Theorems~\ref{thm:switch} and~\ref{thm:general}, Corollary~\ref{cor:unions} and Proposition~\ref{prop:odd}, as in the files \texttt{SwDefs.lean}, \texttt{SwUnions.lean}, \texttt{SwMain.lean} and \texttt{SwProp6.lean} of the repository; \texttt{Hyp}, \texttt{Shatters}, \texttt{ldim} and \texttt{intervals} are those of Figure~\ref{fig:leancode}. \texttt{pathClass a b k n} is the class of sequences $a,f(1),\dots,f(n),b$ with at most $k$ switches, on the $n$ free points. \texttt{cyc k x} is $R_k(x)$: the cycle is read from its fixed point $0$ around and back to it, so its switches are those of $0,f(1),\dots,f(x-1),0$. \texttt{gap k x} is $G_k(x)$, \texttt{switchSeq n k} is $C_{n,k}$, \texttt{unions j m} is $U_j(m)$, and \texttt{Q}, \texttt{v} are $Q_k$, $v_k$ (the value at $k=0$ is not used). The namespace \texttt{ExactLittlestone}, the docstrings and the implicit arguments \texttt{\{a b : Bool\} \{k n d : $\mathbb{N}$\}} of \texttt{shatters\_pathClass\_iff} and \texttt{\{n k : $\mathbb{N}$\}} of the three \texttt{prop6} theorems are omitted, and proofs are replaced by \texttt{\dots}; the last line is the output of the axiom check in \texttt{Check.lean}, which gives the same axioms for the other seven theorems.}
\label{fig:leancode4}
\end{figure}

\begin{figure}[!htbp]
\centering
\includegraphics[width=\textwidth]{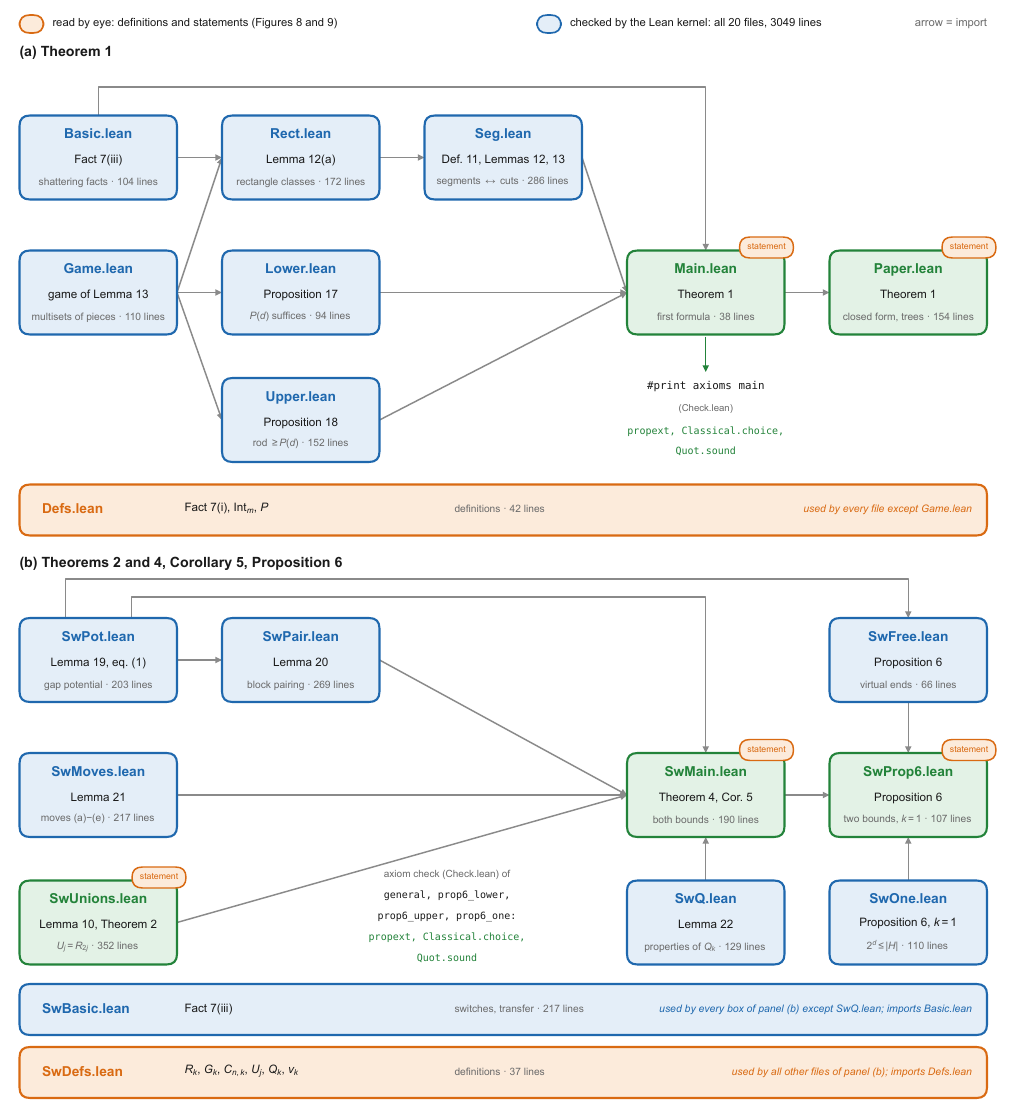}
\caption{Structure of the Lean proofs: (a) Theorem~\ref{thm:main}; (b) Theorems~\ref{thm:switch} and~\ref{thm:general}, the identities of Corollary~\ref{cor:unions} and Proposition~\ref{prop:odd}. Boxes and bands are the $20$ proof files, each labelled with the results of this paper it formalises and its length; arrows are imports, and \texttt{Check.lean}, \texttt{Sanity.lean} and \texttt{SwSanity.lean} are not drawn. \texttt{Defs.lean}, \texttt{SwDefs.lean} and \texttt{SwBasic.lean} are drawn as bands, and their direct imports are not drawn as arrows; \texttt{SwDefs.lean} imports \texttt{Defs.lean}, and \texttt{SwBasic.lean} imports \texttt{SwDefs.lean} and \texttt{Basic.lean}. The definitions in \texttt{Defs.lean} and \texttt{SwDefs.lean} and the statements in \texttt{Main.lean}, \texttt{Paper.lean}, \texttt{SwUnions.lean}, \texttt{SwMain.lean} and \texttt{SwProp6.lean} (Figures~\ref{fig:leancode} and~\ref{fig:leancode4}) are what a reader checks by eye; the Lean kernel checks all $20$ files, and the axiom check lists only the standard axioms of Lean.}
\label{fig:lean}
\end{figure}

\begin{figure}[!htbp]
\centering
\includegraphics[width=\textwidth]{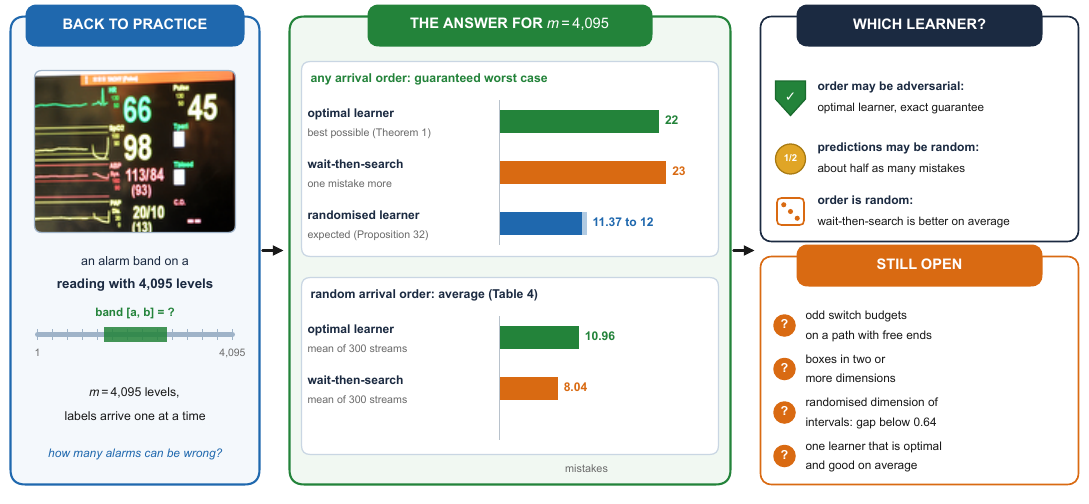}
\caption{What the results give a system that learns an interval rule online, for a reading with $m=4{,}095$ levels. Middle top, any arrival order: the optimal learner can be forced into $\Ldim(\Int_m)=22$ mistakes (Theorem~\ref{thm:main}), the wait-then-search learner into $23$ (Remark~\ref{rem:natural}), and the best randomised learner into an expected number between $R(m+1)=11.37$ and $\tilde\lambda(m+1)=12$ (Proposition~\ref{prop:rlupper}). Middle bottom, random arrival order: mean mistakes from Table~\ref{tab:bench}. Right: which learner suits which setting, and the open problems of Section~\ref{sec:limits}. The photograph is the one of Figure~\ref{fig:motivation}.}
\label{fig:closing}
\end{figure}

Theorems~\ref{thm:main}, \ref{thm:switch} and~\ref{thm:general}, the two identities of Corollary~\ref{cor:unions} and Proposition~\ref{prop:odd} are also verified formally. In the proof assistant Lean~4 \cite{demoura2021lean} with the Mathlib library \cite{mathlib2020lean}, we define the Littlestone dimension by the recursion of Fact~\ref{fact:rec}(i) and prove that it agrees with the definition through complete shattered mistake trees. All proofs are checked by the Lean kernel and use only the standard axioms of Lean.
\begin{itemize}[leftmargin=*]
\item For Theorem~\ref{thm:main}, both formulas are proved for every $m$ (the second for $m\ge3$). The proof takes a shorter route than Sections~\ref{sec:cut} and~\ref{sec:tree}: the positive branch of a query is treated as a class of intervals with two ranges of endpoints (proof of Lemma~\ref{lem:branch}), and the lower bound on the rod uses only the first three cuts (the paragraph after Proposition~\ref{prop:lower}).
\item Theorem~\ref{thm:general} is proved for every $k\ge1$ and $x\ge1$, and in the form with any fixed end labels stated at the start of Section~\ref{sec:general}. The proof follows Section~\ref{sec:general}: the potential of Lemma~\ref{lem:pot} (for two labelled ends), the moves of Lemma~\ref{lem:moves} and Lemma~\ref{lem:Q}. Lemma~\ref{lem:pair} is proved in the tree form given after its proof, so no learner is formalised.
\item Theorem~\ref{thm:switch} follows from Lemma~\ref{lem:cyc}, and $U_j(m)$ is shown to be equal to $R_{2j}(m+1)$ as a class on $[m]$; with Theorem~\ref{thm:general} this gives the identities of Corollary~\ref{cor:unions}.
\item Proposition~\ref{prop:odd} is proved for every $n\ge1$ and odd $k$. The bound $\lfloor(n+k-1)/2\rfloor$ follows from the potential of Lemma~\ref{lem:pot} for two labelled ends after adding virtual end labels, and the case $k=1$ uses the counting form given at the end of the proof of Proposition~\ref{prop:odd}.
\end{itemize}
The link between the Littlestone dimension and mistake bounds \cite{littlestone1988learning}, Corollary~\ref{cor:asym}, the estimates of Corollary~\ref{cor:unions} and Sections~\ref{sec:remarks} and~\ref{sec:random} are not part of the formal proof. Figures~\ref{fig:leancode} and~\ref{fig:leancode4} give the definitions and statements. A reader needs to compare only these lines, together with the Mathlib definitions they use (finite sets, \texttt{Finset.powerset}, \texttt{List.ofFn}, \texttt{Nat.findGreatest}, \texttt{Nat.log}, \texttt{IsGreatest}, \texttt{Even} and \texttt{Odd}; subtraction and division on $\mathbb{N}$ are truncated), with Fact~\ref{fact:rec} and the statements of these results. Figure~\ref{fig:lean} shows how the rest of the proof is organised. The Lean files are in the same repository.

\section{Conclusion}\label{sec:open}

For intervals on $m$ points, the optimal mistake bound is the largest $d$ with $P(d)\le m+1$, where $P(d)$ alternates between $5\cdot2^{t-1}$ and $7\cdot2^{t-2}$, so for $m\ge3$ the dimension equals $2\lg(m+1)$ minus a quantity between $1.615$ and $2.644$. Unions of $j$ intervals and switch-bounded sequences follow one three-regime law for $Q_k$, which we prove by matching a gap potential and a block-pairing learner against adversary moves and a refinement step. A second proof for intervals, through the cutting game and admissible trees, describes every winning cut sequence of the adversary and carries over to randomised learners. Theorems~\ref{thm:main}, \ref{thm:switch} and~\ref{thm:general} and Proposition~\ref{prop:odd} are also verified formally in Lean~4.

Exact values settle two questions that order-of-growth bounds leave open. The natural learner, which waits for a positive example and then binary-searches both endpoints, is optimal for some domain sizes and makes one mistake more than necessary for others ($50{,}847$ of the $m<10^5$), so an optimal learner must sometimes predict a positive label before it has seen one. Counting is off as well: the bound $\lgf{|\Int_m|}$ exceeds the dimension by one for $62{,}063$ of these $m$. For randomised learners the exact dimension of thresholds is half the interpolated logarithm, and for intervals it lies between $\tilde\lambda(m+1)-c^\ast$ and $\tilde\lambda(m+1)$, where $\tilde\lambda$ is the interpolated logarithm and $c^\ast<0.64$, so it grows at half the deterministic rate. For a system that learns interval or switch-based rules online, the results give the exact number of mistakes it can guarantee, and the common wait-then-search strategy falls short of that number. The potential-guided learner of Lemma~\ref{lem:pot}, combined with the block-pairing learner of Lemma~\ref{lem:pair} as in the upper-bound proof of Theorem~\ref{thm:general}, attains it at $O(\log m)$ cost per round (Section~\ref{sec:learner}). In the random-stream experiments of Table~\ref{tab:bench}, however, it makes more mistakes on average than the natural learner. Figure~\ref{fig:closing} sets these numbers side by side for a reading with $m=4{,}095$ levels.

\subsection{Limitations and open problems}\label{sec:limits}

The exact results cover one-dimensional domains. For odd $k$ on a path with free ends, Proposition~\ref{prop:odd} gives matching bounds only for $k=1$; equality for $k=3,5,7$ rests on exhaustive computation for small $n$. For randomised learners the dimension of intervals is determined only to within $c^\ast<0.64$, and the lower bound is checked to be exact for $m\le70$ and on all positions of total measure at most $30$. The learner of Section~\ref{sec:learner} attains the worst-case bound, but in the random-stream experiments of Table~\ref{tab:bench} the natural learner makes fewer mistakes on average. These gaps lead to the following problems.
\begin{enumerate}[label=(\arabic*)]
\item Prove $\Ldim(C_{n,k})=1+v_k(n)$ for odd $k\ge3$ (Proposition~\ref{prop:odd}). The obstacle is the upper bound for a first query in the interior of the path: the pairing argument of Lemma~\ref{lem:pair} needs a block structure with $\lfloor n/2\rfloor$ blocks, which free ends do not always allow.
\item Determine the exact Littlestone dimension of axis-parallel boxes in $\{1,\dots,n\}^d$ for $d\ge2$, where only $\Theta(d\log n)$ is known \cite{maass1994algorithms,chen1992line,chen1994line}.
\item Prove Conjecture~\ref{conj:rl}, i.e.\ that the greedy value is a supersolution of the discounted cutting game; then determine the randomised Littlestone dimension of $U_j(m)$ for $j\ge2$.
\item For every $m$, find a learner for $\Int_m$ that makes at most $\Ldim(\Int_m)$ mistakes against every adversary and, on a uniformly random target presented in uniformly random order, no more mistakes on average than the natural learner. Where $N(m)=\Ldim(\Int_m)$ the natural learner itself qualifies, so the question concerns the remaining $m$.
\end{enumerate}

\section*{Data and code availability}
No datasets were used. The code, written in Python with only the standard library, is publicly available on GitHub at \url{https://github.com/researchbackup27-sudo/exact-littlestone-intervals} under the MIT license. It contains a reference implementation of the objects and procedures of this paper (the thresholds $P$ and $Q_k$, the cutting game and the tree minimisation, the gap-state dynamic programs, the bounds of Section~\ref{sec:general} and the randomised recursion) and of the optimal learner built in the upper-bound proof of Theorem~\ref{thm:general}. A demonstration script reruns every computational check in Table~\ref{tab:verification} and plays this learner against an exhaustive adaptive adversary on small domains. The repository also has a script that counts the mistakes of the optimal and natural learners on random targets presented in random order, and the Lean~4 formal proofs of Theorems~\ref{thm:main}, \ref{thm:switch} and~\ref{thm:general} and Proposition~\ref{prop:odd}. On publication, the repository will be archived on Zenodo with a DOI.

\section*{Declaration of competing interest}
The authors declare that they have no known competing financial interests or personal relationships that could have appeared to influence the work reported in this paper.

\section*{Author contributions}
\textbf{Miit Daga:} Conceptualization, Methodology, Software, Validation, Formal analysis, Investigation, Data curation, Visualization, Writing -- original draft. \textbf{Swarna Priya Ramu:} Conceptualization, Validation, Writing -- review \& editing, Supervision, Project administration.

\bibliographystyle{plainnat}
\bibliography{references}

\end{document}